\documentclass[lettersize,journal]{IEEEtran}

\usepackage{algorithm}
\usepackage{algorithmic}

\usepackage{amsthm}
\usepackage{amsmath,amssymb,mathtools}
\usepackage{makecell}
\usepackage{booktabs}
\usepackage{footnote}
\usepackage{cite}
\usepackage[switch]{lineno}  %

\newtheorem{theorem}{Theorem}

\newtheorem{lemma}{Lemma}
\newtheorem{corollary}{Corollary}
\newtheorem{assumption}{Assumption}
\newtheorem{definition}{Definition}

\newtheorem{remark}{Remark}
\newcounter{MYtempeqncnt}
\usepackage{ifpdf}
\usepackage[flushleft]{threeparttable}
\usepackage{makecell}
\usepackage{multirow}
\usepackage{bbding}

\ifCLASSINFOpdf
\else
\fi
\begin{document}
%
\title{Communication-Efficient Distributed Learning with Local Immediate Error Compensation}
%
%
%

\author{Yifei~Cheng, Li~Shen, Linli~Xu, Xun~Qian, Shiwei~Wu, Yiming~Zhou, Tie~Zhang, \\Dacheng~Tao,~\IEEEmembership{Fellow,~IEEE}, Enhong~Chen,~\IEEEmembership{Fellow,~IEEE}

\IEEEcompsocitemizethanks{
\IEEEcompsocthanksitem Y. Cheng, S. Wu, Y. Zhou, T. Zhang, L. Xu and E. Chen are with the Anhui Province Key Lab of Big Data Analysis and Application, State Key Laboratory of Cognitive Intelligence, School of Data Science and School of Computer Science, University of Science and Technology of China. E-mail: \{chengyif, dwustc, zym2019, tiezhang\}@mail.ustc.edu.cn, \{linlixu, cheneh\}@ustc.edu.cn.\protect
\IEEEcompsocthanksitem L. Shen and D. Tao are with the JD Explore Academy. E-mail: \{mathshenli, dacheng.tao\}@gmail.com.\protect
\IEEEcompsocthanksitem X. Qian is with the Shanghai Artificial Intelligence Laboratory. E-mail: qianxun@pjlab.org.cn \protect
\IEEEcompsocthanksitem Corresponding authors: Li Shen, Enhong Chen.}
}

%
%

\markboth{Journal of \LaTeX\ Class Files,~Vol.~14, No.~8, August~2015}%
{Shell \MakeLowercase{\textit{et al.}}: Bare Advanced Demo of IEEEtran.cls for IEEE Computer Society Journals}
%



\IEEEtitleabstractindextext{%

\begin{abstract}
Gradient compression with error compensation
has attracted significant attention with the target of reducing the heavy communication overhead in distributed learning. However, existing  compression methods either perform only unidirectional compression in one iteration with higher communication cost, or bidirectional compression with slower convergence rate. In this work, we propose the Local Immediate Error Compensated SGD (LIEC-SGD) optimization algorithm to break the above bottlenecks based on bidirectional compression and carefully designed compensation approaches. Specifically, the bidirectional compression technique is to reduce the communication cost, and the compensation technique compensates the local compression error to the model update immediately while only maintaining the global error variable on the server throughout the iterations to boost its efficacy.  Theoretically, we prove that LIEC-SGD is superior to previous works in either the convergence rate or the communication cost, which indicates that LIEC-SGD could inherit the dual advantages from unidirectional compression and bidirectional compression. Finally, experiments of training deep neural networks validate the effectiveness of the proposed LIEC-SGD algorithm.
\end{abstract}

}

\maketitle

\IEEEdisplaynontitleabstractindextext

%
\IEEEpeerreviewmaketitle

\section{Introduction}
\IEEEPARstart{S}{upported} by the powerful hardware resources, the data and model scale in machine learning increase rapidly, followed by the significant challenges imposed to efficient large-scale machine learning. To tackle the challenges, distributed learning across multiple workers becomes an important and successful principle. A conventional framework for distributed learning is the centralized worker-server architecture where the whole dataset is split and distributed over the workers~\cite{li2014scaling,li2014communication}. In this architecture, $N$ workers are coordinated together 
to train a model by minimizing the following loss function:
\begin{equation}\label{problem}
    f(x) = \frac{1}{N} \sum_{i=1}^N \mathbb{E}_{\xi_i \sim \mathcal{D}_i} f_i(x,\xi_i), x\in R^d
\end{equation} 
where $\mathcal{D}_i$ and $f_i(x,\xi_i)$ denote the data distribution and loss function on the $i$-th worker respectively.

A standard optimization algorithm in distributed learning is Parallel Stochastic Gradient Descent (P-SGD), which runs SGD in parallel across all workers. In every iteration, the workers calculate the stochastic gradients based on the local data, then the server gathers these local gradients and sends the averaged gradient back to the workers for updating the model. However, P-SGD is limited significantly in large-scale machine learning due to the heavy communication overhead caused by the frequent high-dimensional gradients exchange, especially in cases with poor network conditions. As a consequence, the high communication cost becomes a bottleneck which constrains the performance of distributed learning markedly~\cite{you2018imagenet}.

Gradient compression is an effective strategy to overcome the communication bottleneck mentioned above by transmitting very small size vectors~\cite{tang2020communication, xu2021grace}. Quantization~\cite{seide20141,bernstein2018signsgd,wen2017terngrad,alistarh2017qsgd} and sparsification methods~\cite{strom2015scalable,aji2017sparse,lin2018deep,stich2018sparsified} are proposed to produce a representation of the gradients with less precision compared to the full-precision gradients in P-SGD.

Nevertheless, too aggressive compression strategies may influence the convergence performance~\cite{seide20141,stich2018sparsified}. Thus
some works consider compensating the errors generated by the compression to the gradients computed later. Specifically,~\cite{seide20141,wu2018error} couple the quantization compression with the error-compensation framework without deterioration of performance in distributed training. Sparsification compression with error compensation is extensively studied in~\cite{stich2018sparsified}, where 
parallel memory SGD (\textbf{MEM-SGD}) sparsifies the gradients sent from the workers to the server. The above methods all use unidirectional compression, i.e. the uplink. In order to more adequately reduce the communication cost, downlink compression should also be implemented~\cite{philippenko2020bidirectional}. DoubleSqueeze~\cite{tang2019doublesqueeze} applies the bidirectional compression, i.e. compressing the gradients sent in both directions. However, according to the theoretical analysis in ~\cite{stich2018sparsified,zheng2019communication,xu2021step,sahu2021rethinking}, the algorithm with bidirectional compression 
converges more slowly, since it suffers more from the gradients remained in the error variables than the algorithm with unidirectional compression. Neolithic \cite{huang2022lower} adopts recursive compression on both the worker and server side, and achieves the nearly optimal convergence rate. Nevertheless, multiple rounds of compression and communication are not efficient in practical implementation. Thus, it is not clear whether the bidirectional compression and fast convergence rate can be achieved simultaneously in an efficient way. In this paper, we propose an algorithm with a new compensation strategy and solve the issues mentioned above. Our main contributions are summarized as follows:

\begin{figure}[!t]
\centering
\includegraphics[scale=0.36]{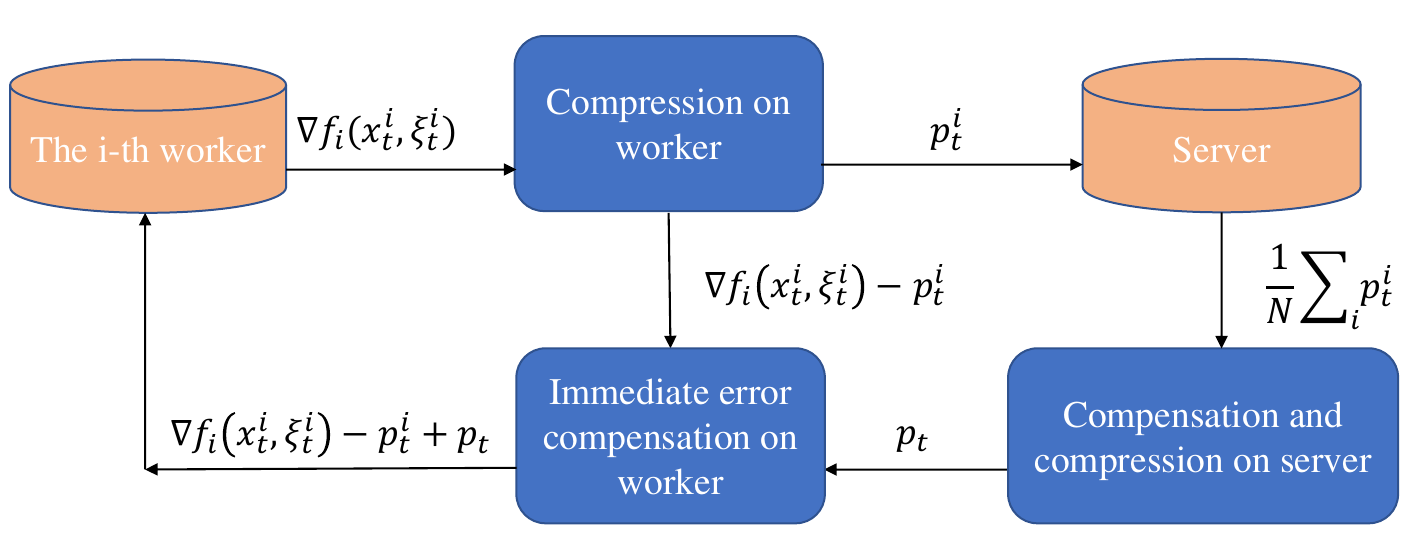}
\caption{Workflow of LIEC-SGD. When local immediate error-compensation is implemented, the stochastic gradient is compressed and the compression error is cached on the worker. The server gathers and averages the compressed gradients and compensates the global error to it. The result is then compressed by the server and broadcast to the worker. Each worker compensates the cached compression error to the returned value to update the model.}
\label{flow}
\end{figure}

\textbf{Novel Distributed Algorithm.} We propose an algorithm LIEC-SGD (see Figure \ref{flow}) which adopts a carefully designed Local Immediate Error Compensation framework and bidirectional compression. The key of the algorithm is that each worker directly compresses the stochastic gradient, then the compression error would be compensated to the gradients returned from the server to update the model. Consequently, the local compression error would be involved in model update immediately instead of remaining in the error as traditional error-compensation methods. Considering the discrepancies among local models, we periodically average the model parameters and clear the error on the server.

\textbf{Theoretically Analysis.} In the comparison with traditional error-compensation algorithms, our theoretical results show that when adopting the $\delta$-contraction operator, LIEC-SGD achieves better convergence rate than those with bidirectional compression, and the same as those with unidirectional compression. In addition, we relax the gradient bounded assumption (Assumption \ref{assu3}) which is used in most previous works, in our analysis (\textbf{Case 1} in section III). We summarize these theoretical results in Table~\ref{table1}.

\textbf{Empirical Verification.} We conduct sufficient experiments of training deep neural networks. The experimental results indicate that LIEC-SGD reduces the norm of error more effectively and achieves the best testing performance among all the baselines. Benefited from bidirectional gradient compression, the time cost per epoch of our proposed algorithm is also at the lowest level. The empirical studies verify the superiority of the proposed algorithm.

\textbf{Notations.} We use the following notations throughout this work: $\|\cdot\|$ denotes the $l_2$ norm of a vector; $\nabla f(x)$ represents the gradient of $f(x)$; $x^*$ denotes the optimal solution of (\ref{problem}); $T$ indicates the total number of iterations; $N$ indicates the total number of workers; $[N]$ represents the set $\{1,2,...,N\}$.

\begin{table*}[t]
\renewcommand{\arraystretch}{0.4}
    \caption{Comparison of Different Distributed Optimization Algorithms for Non-convex Problems.}
    \centering
    \begin{threeparttable}
    \begin{tabular}{ccccc}
        \toprule \\
        Algorithm & Convergence rate & Compression & \makecell[c]{Threshold of $T$ to\\achieve linear speedup\tnote{a}} & \makecell[c]{No need of extra\\assumption} \\
        \toprule \\
        P-SGD & $O\bigg(\frac{1}{\sqrt{NT}}+\frac{1}{T}\bigg)$ & No & $O(N)$ & - \\
        \hline \\
        MEM-SGD~\cite{stich2018sparsified} & $O\bigg(\frac{1}{\sqrt{NT}}+\frac{1}{\delta^{2/3} T^{2/3}}+\frac{1}{T}\bigg)$ & Unidirectional & $O(N^3/\delta^4)$ & \XSolidBrush \\
        \hline \\
        DoubleSqueeze~\cite{tang2019doublesqueeze} & $O\bigg(\frac{1}{\sqrt{NT}}+\frac{1}{\delta^{4/3} T^{2/3}}+\frac{1}{T}\bigg)$ & Bidirectional & $O(N^3/\delta^8)$ & \XSolidBrush \\
        \hline \\
        Step-Ahead~\cite{xu2021step} & $O\bigg(\frac{1}{\sqrt{NT}}+\frac{1}{\delta^{4/3} T^{2/3}}+\frac{1}{T}\bigg)$ & Bidirectional & $O(N^3/\delta^8)$ & \XSolidBrush \\
        \hline \\
        CSER~\cite{xie2020cser} & $O\bigg(\frac{1}{\sqrt{NT}}+\frac{H^{2/3}}{\delta_1^{2/3} T^{2/3}}+\frac{1}{T}\bigg)$\tnote{c} & Unidirectional & $O(N^3 H^4/\delta_1^{4})$ & \XSolidBrush \\
        \hline \\
        Neolithic~\cite{huang2022lower} & $\Tilde{O}\bigg(\frac{1}{\sqrt{NT}} + \frac{1}{\delta T}\bigg)$ & Bidirectional & $\Tilde{O}(N/\delta^2)$  & \Checkmark \\
        \hline \\
        This work & $O\bigg(\frac{1}{\sqrt{NT}}+\frac{1}{\delta^{2/3} T^{2/3}}+\frac{1}{T}\bigg)$ & Bidirectional & $O(N^3/\delta^4)$ & \Checkmark \\
        \toprule \\
    \end{tabular}
    \begin{tablenotes}
    \item[a] This denotes the number of iterations beyond which the algorithm achieves the linear speedup.
    \item[b] This represents that whether the theoretical result needs the assumption that the stochastic gradient is bounded: $\forall i \in [N]$, there exists a constant $M$ such that $\mathbb{E}_{\xi \sim \mathcal{D}_i} \|\nabla f_i(x, \xi)\|^2 \leq M^2$, or the error is bounded \cite{tang2019doublesqueeze}: $\mathbb{E}\|\delta_t^i\|\leq \frac{\epsilon}{2}, \mathbb{E}\|\delta_t\|\leq \frac{\epsilon}{2}$.
    \item[c] $H$ denotes the period to reset the error. 
    \end{tablenotes}
    \end{threeparttable}
    \label{table1}
\end{table*}
\section{Related Work}
\subsection{Distributed Training}
Distributed training is widely applied to accelerate large-scale machine learning, especially deep learning models~\cite{chen2016revisiting,jia2019beyond}. It has been maturely developed in popular machine learning applications~\cite{abadi2016tensorflow,paszke2019pytorch}. There are two common patterns of distributed learning, model parallelism and data parallelism~\cite{dean2012large}. In model parallelism, each worker is responsible for a part of model parameters while in data parallelism, each worker is accessible to only a part of data but all the model parameters~\cite{krizhevsky2017imagenet}. In this paper, we focus on data parallelism. 

As a representative distributed optimization method, P-SGD is extensively used for distributed training. It has been proved that P-SGD achieves linear speedup with respect to the number of workers~\cite{recht2011hogwild,dekel2012optimal,lian2015asynchronous}. However, the communication overhead caused by the gradients exchange between workers becomes the bottleneck of the actual performance, especially for large-scale machine learning tasks where the model size is huge and training time is long~\cite{li2014communication,lin2018deep}.
\vspace{-0.2cm}
\subsection{Gradient Compression}
Compressing the gradients is considered as an efficient way to reduce the communication cost of distributed training by approximating the accurate values of gradient elements with very few signs or digits. As a mainstream of compression principles, quantization methods are studied in many works recently.~\cite{seide20141} proposes to quantize the gradients to only 1-bit.~\cite{wen2017terngrad} suggests representing the gradients with three numerical levels and proves the convergence of the algorithm. QSGD is proposed to allow the workers to adjust the number of bits transferred, thus yielding a trade-off between the convergence rate and communication cost~\cite{alistarh2017qsgd}.~\cite{jiang2018sketchml} studies the compression framework for sparse vectors specially. The quantization operator can also be coupled with Local-SGD to further reduce the communication cost~\cite{jiang2018linear}.

Another direction of gradient compression is sparsification that drops a mass of unnecessary elements. Compared to quantization, sparsification is more aggressive while enjoying higher compression ratio~\cite{yang2021fastsgd}. \cite{wangni2018gradient} proposes to drop out the elements randomly according to the distribution that achieves the lowest sparsity given the fixed variance budget. In~\cite{aji2017sparse,stich2018sparsified,shi2019understanding}, top-$k$ sparsification is considered as the compression operator.~\cite{strom2015scalable,chen2018adacomp,yang2021fastsgd,sahu2021rethinking,m2021efficient} use the threshold filter to retain the elements according to their values or norm.

In addition, these operators could also be applied to compress the gradient differences \cite{mishchenko2019distributed}. Based on this, \cite{liu2020double} considers to compress the model information in the server-to-worker direction to realize bidirectional compression. \cite{philippenko2021preserved} adopts a memory process to focus on the preserved central model, while \cite{gruntkowska2023ef21} acts as a framework to decouple and correct the error incurred by the compression. However, compressing gradient differences needs to keep track of the historical
information, thus increasing the memory cost. More importantly, most of the theoretical results in these works rely on the unbiased compression which restricts the choices of compressors, thus we do not list them in Table \ref{table1}. We compare the representative of them with our approach in the part of the experiment.

Note that the compression operators mentioned above would introduce extra computation overhead, which sometimes even surpasses the communication cost saved~\cite{eghlidi2020sparse,m2021efficient}. As a result, the wall-clock time of the training process of some methods may be longer than that of P-SGD. Thus, the training time speedup is an important metric for evaluating the performance of distributed algorithms.

\subsection{Error-compensation Framework}
The error-compensation framework is introduced in~\cite{seide20141} to compensate the error of 1-bit quantization to the next mini-batch gradient. Further, EF-SignSGD which reforms SignSGD~\cite{bernstein2018signsgd} and combines it with error-feedback is proposed in~\cite{karimireddy2019error} and proved to converge for non-convex problems. \cite{wu2018error} introduces error compensated quantization and proves its convergence for quadratic optimization. In~\cite{stich2018sparsified,alistarh2018convergence,shi2019understanding,stich2020error,horvath2021better,xu2022detached}, applying the sparsification operator with error-compensation is sufficiently studied with the convergence guaranteed. \cite{tang2019doublesqueeze,zheng2019communication} consider implementing compression bidirectionally to reduce communication cost and prove the linear speedup of the algorithm. \cite{huang2022lower} proposes Fast Compressed Communication (FCC) which recursively compresses the residuals on both the workers and server. The vanishing error benefited by FCC helps the algorithm to almost reach the lower bound. \cite{basu2019qsparse,gao2021convergence} show that gradient compression with error-compensation is compatible well with Local-SGD both theoretically and empirically. In addition, the low-rank compressor is proposed to reduce the communication cost with error-compensation~\cite{vogels2019powersgd}. Though~\cite{richtarik2021ef21} achieves fast convergence rate, it requires the global model broadcast per iteration. Recently, \cite{chen2021quantized} 
applies the error-compensation framework to Adam, but the method merely converges to the neighborhood of the stationary point. Finally, \cite{gorbunov2020linearly} makes an analysis for SGD with error-compensation under a unified framework. 
\section{Methodology}

\subsection{Problem Setup}
\subsubsection{Definition}
In the error-compensation framework, the vector or the tensor sent between the workers and the parameter server is commonly compressed by an $\delta$-contraction operator which is defined as follows:
\begin{definition}[$\delta$-contraction operator]\label{def1}
A compression operator $\mathcal{C}_\delta: \mathbb{R}^d \rightarrow \mathbb{R}^d$ is a $\delta$-contraction operator if it satisfies
\begin{equation}
    \mathbb{E}\|x-\mathcal{C}_\delta(x)\|^2 \leq (1-\delta) \|x\|^2, \forall x\in R^d. \nonumber
\end{equation}
\end{definition}
The following compression operators are examples of $\delta$-contraction operators~\cite{stich2018sparsified,qian2020error,karimireddy2019error,zheng2019communication}:
\begin{definition}[top-k]\label{def2}
For a parameter $1\leq k \leq d$, the operator top-k: $\mathbb{R}^d \rightarrow \mathbb{R}^d$ is defined as 
\begin{equation}
    (\mathcal{C}(x))_{\pi(i)} =
    \begin{cases}
    (x)_{\pi(i)},& if\ i\leq k, \\
    0,& otherwise, \nonumber
    \end{cases}
\end{equation}
where $\pi$ is a permutation of $\{1,2,...,d\}$ such that $(|x|)_{\pi(i)} \geq (|x|)_{\pi(i+1)}$ for $i=1,2,...,d-1$.
\end{definition}

\begin{definition}[random-k]\label{def3}
For a parameter $1\leq k \leq d$, the operator random-k: $\mathbb{R}^d \rightarrow \mathbb{R}^d$ is defined as 
\begin{equation}
    (\mathcal{C}(x))_{(i)} =
    \begin{cases}
    (x)_{(i)},& if\ i\in \mathcal{S}, \\
    0,& otherwise, \nonumber
    \end{cases}
\end{equation}
where $\mathcal{S}$ is sampled from $k$-element subsets of [d] uniformly.
\end{definition}

\begin{definition}[SignSGD]\label{def4}
The operator SignSGD: $\mathbb{R}^d \rightarrow \mathbb{R}^d$ is defined as 
\begin{equation}
    (\mathcal{C}(x))_{(i)} = \|x\|_1 / d \cdot sign(x) \nonumber
\end{equation}
\end{definition}

\begin{definition}[Blockwise-SignSGD]\label{def5}
The operator Blockwise-SignSGD: $\mathbb{R}^d \rightarrow \mathbb{R}^d$ is defined as 
\begin{equation}
    (\mathcal{C}(x))_{(i)} = [\|x_1\|_1 / d_1 \cdot sign(x_1), ... , \|x_k\|_1 / d_k \cdot sign(x_k)], \nonumber
\end{equation}
where $x_i, i \in [k]$ represents a subset of elements in $x$, and $d_i$ denotes its dimension.
\end{definition}

Obviously, the above random-k operator is of $\delta$-contraction operator with $\delta=\frac{k}{d}$ and the top-k operator is of $\delta$-contraction operator with $1>\delta \geq \frac{k}{d}$~\cite{stich2018sparsified,sahu2021rethinking}. Thus, for these two operators, $\delta$ reflects the upper bound of the sparsity of the vectors after being compressed by them.~\cite{karimireddy2019error} reforms the "SignSGD" in \cite{bernstein2018signsgd} by considering the magnitude of the gradient. Blockwise-SignSGD divides the gradient into several blocks to implement SignSGD thereby making the compression more granular. Obviously, Blockwise-SignSGD reduces to SignSGD when the number of blocks equals to 1. As listed in~\cite{zheng2019communication}, the compression ratio is 32x for SignSGD since the signal of a vector takes only 1-bit, and nearly 32x for Blockwise-SGD. 

\subsubsection{Assumptions} 
Throughout this paper, we make the following assumptions which are commonly used in previous works \cite{stich2018sparsified,zheng2019communication,basu2019qsparse,xie2020cser} for our theoretical analysis. 
\begin{assumption}[Lipschitzian gradient]\label{assu1}
    For each $i \in [N]$, $f_i(x)$ is with $L$-Lipschitzian gradient:
    \begin{equation}
        \| \nabla f_i(x) - \nabla f_i(y) \| \leq L \|x - y \|, \forall x, y \in R^d.
    \nonumber
    \end{equation}
\end{assumption}
\begin{assumption}[Bounded variance of stochastic gradients]\label{assu2}
  $\forall i \in [N]$, there exists a constant $\sigma$ such that 
  \begin{equation}
      \mathbb{E}_{\xi \sim \mathcal{D}_i} \| \nabla f_i(x, \xi) - \nabla f_i (x) \|^2 \leq \sigma^2, \forall x \in R^d.
      \nonumber
  \end{equation}
\end{assumption}

\begin{assumption}[Bounded second moments]\label{assu3}
  $\forall i \in [N]$, there exists a constant $M$ such that
  \begin{equation}
      \mathbb{E}_{\xi \sim \mathcal{D}_i} \|\nabla f_i(x, \xi)\|^2 \leq M^2, \forall x \in R^d. \nonumber
  \end{equation}
\end{assumption}

We then make the following assumption for some special cases of the $\delta$-compression operators.
\begin{assumption}\label{assu4}
For the operator $\mathcal{C}_\delta$ and $\forall x \in R^d$,
\begin{equation}
\mathbb{E}[\mathcal{C}_\delta(x)] = \delta x. \nonumber
\end{equation}
Random-k operator obviously satisfies this assumption with $\delta = k/d$~\cite{qian2020error}.
\end{assumption}

In addition, we provide the \textit{strong growth condition} (SGC) which is an interpolation-like condition.
\begin{assumption}\label{assu5}
$f(x)$ satisfy SGC with the constant $\rho$ if
\begin{equation}
    \mathbb{E}\|f_i(x)\|^2 \leq \rho \|f(x)\|^2, \forall x \in R^d. \nonumber
\end{equation}
\end{assumption}

\subsection{LIEC-SGD Algorithm}

We propose the Local Immediate Error Compensation SGD (LIEC-SGD) as described in Algorithm~\ref{LIEC}. We use the notations $p_t^i$ and $p_t$ to represent the vector sent from $i$-th worker to server and server to $i$-th worker respectively.
\begin{algorithm}[tb]
  \caption{LIEC-SGD}\label{LIEC}
  \begin{algorithmic}[1]
  \REQUIRE Initialize $x_0^i=x_0$, learning rate $\eta$, error $e_0^i=e_0=0$, number of iterations $T$.
  \FOR{$t = 0, ..., T-1$}
    \STATE \textbf{Worker $i$:}
    \STATE Computes local gradient $\nabla f_i(x_t^i, \xi_t^i)$ 
    \IF {$(t+1) \% \lfloor \frac{1}{\delta} \rfloor = 0$}
        \STATE Sends $p_t^i = \nabla f_i(x_t^i, \xi_t^i)$ and $x_t^i$ to the server
    \ELSE
        \STATE Compresses $\nabla f_i(x_t^i, \xi_t^i)$ with operator $\mathcal{C}_\delta: p_t^i = \mathcal{C}_\delta (\nabla f_i(x_t^i, \xi_t^i))$ and sends $p_t^i$ to the server
    \ENDIF
    \STATE \textbf{Server:}
    \STATE Compensates the average of $p_t^i$'s with the global error $v_t=e_t+\frac{1}{N}\sum_{i=1}^N p_t^i$
    \IF {$(t+1) \% \lfloor \frac{1}{\delta} \rfloor = 0$}
        \STATE Sends $p_t = v_t$ and $\frac{1}{N}\sum_{i=1}^N x_t^i$ to the workers
    \ELSE
        \STATE Compresses $v_t$ with operator $\mathcal{C}_\delta: p_t = \mathcal{C}_\delta (v_t)$ and sends $p_t$ to the workers
    \ENDIF
    \STATE Updates the global error $e_{t+1} = v_t-p_t$
    \STATE \textbf{Worker $i$:}
    \STATE Updates the local model: 
    \IF {$(t+1) \% \lfloor \frac{1}{\delta} \rfloor = 0$}
        \STATE $x_{t+1}^i = \frac{1}{N}\sum_{i=1}^N x_t^i-\eta (p_t-p_t^i+\nabla f_i(x_t^i, \xi_t^i))$
    \ELSE
        \STATE $x_{t+1}^i = x_t^i-\eta (p_t-p_t^i+\nabla f_i(x_t^i, \xi_t^i))$
    \ENDIF
  \ENDFOR
    \ENSURE $\bar{x}_T = \frac{1}{N} \sum_{i=1}^N x_T^i$
  \end{algorithmic}
\end{algorithm}

In the $t$-th iteration with $(t+1) \% \lfloor \frac{1}{\delta} \rfloor \neq 0$, each worker first computes and compresses the local gradient (line 7). The compression error is cached on the worker for the moment. The server calculates the average of the compressed gradients and compensates the global error to it (line 10). The $\delta$-contraction operator is applied to obtain the compressed vector which is sent back to the workers (line 14). The global error is then updated (line 16). The next step is the key of our algorithm (line 22), where the update of each worker is designed as
\begin{equation}
    x_{t+1}^i = x_t^i - \eta(p_t-p_t^i+\nabla f_i(x_t^i, \xi_t^i)). \nonumber
\end{equation}

Specifically, each worker compensates the cached local error $\nabla f_i(x_t^i, \xi_t^i)-p_t^i$ to the returned value $p_t$ immediately and updates the local model $x_t^i$ with the result. 
For better illustration, we show the workflow of these iterations in Figure~\ref{flow}.

Otherwise when $(t+1) \% \lfloor \frac{1}{\delta} \rfloor = 0$, the full gradients and the local model parameters are transferred to the server (line 5). The server performs the same compensation step (line 10) and sends the result back to the workers along with the averaged model parameters (line 12). The averaged model would unify the local models on all workers to be the same. Notice that no compression is performed in this iteration, i.e., the global error is compensated to the averaged gradients and sent to the workers integrally. Thus this step does not produce the error, the update of global error sets it to zero (line 16). Each worker updates the local model based on the averaged model parameters (line 20) as
\begin{eqnarray}
    x_{t+1}^i &=& \frac{1}{N}\sum_{i=1}^N x_t^i - \eta(p_t-p_t^i+\nabla f_i(x_t^i, \xi_t^i)) \nonumber \\
    &=& \frac{1}{N}\sum_{i=1}^N x_t^i - \eta(\frac{1}{N}\sum_{i=1}^N \nabla f_i(x_t^i, \xi_t^i) + e_t). \nonumber
\end{eqnarray}
It is worth noting that model averaging and gradient communication could be conducted in parallel, thus incurring little extra time cost. 
In addition, the number of the iterations 
satisfying $(t+1) \% \lfloor \frac{1}{\delta} \rfloor \neq 0$ is $(1-1/\lfloor \frac{1}{\delta} \rfloor)T \leq (1-\delta)T$, thus the communication complexity of these iterations is $O(\frac{k}{d}(1-\delta)T)$. And the communication complexity of the iterations 
satisfying $(t+1) \% \lfloor \frac{1}{\delta} \rfloor = 0$ is $O(T/\lfloor \frac{1}{\delta} \rfloor)$.

\textbf{Discussion.} In traditional error-compensation algorithms (for instance MEM-SGD and DoubleSqueeze), the updated values in each iteration contain the gradients computed previously which are dropped by the compression operator. Compared to P-SGD, these gradient elements are used to update the model later than when they are calculated as they are remained in the error variables (for instance, $\frac{1}{N}\sum_{i=1}^N e_t^i$ for MEM-SGD and $\frac{1}{N}\sum_{i=1}^N e_t^i + e_t$ for DoubleSqueeze). As pointed out in \cite{xu2021step}, the stale pattern is similar to asynchronous SGD which updates the model with stale gradients~\cite{lian2015asynchronous,alistarh2018convergence}. This phenomenon of remaining gradients 
has also been discussed in~\cite{xu2021step,sahu2021rethinking}. Particularly, the first work names it "gradient mismatch" and proposes the step ahead technique to alleviate this problem. However, there still exist error variables on both the workers and the parameter server. \\
As a comparison, LIEC-SGD adopts local immediate error-compensation on the worker side in most iterations (when $(t+1) \% \lfloor \frac{1}{\delta} \rfloor \neq 0$). The local error caused by the gradient compression is arranged to update the model without any delay. In the rest of the iterations (when $(t+1) \% \lfloor \frac{1}{\delta} \rfloor = 0$), besides averaging the model to eliminate the discrepancies among local models, we force the gradient elements remained in the global error to be sent to update the model immediately. Moreover, clearing the error to zero could also prevent it from diverging too fast which is similar to error averaging in~\cite{xu2021step}. Overall, our operations on the error reduce the length of the delay and allow the gradients in the error to participate in the model updates in advance compared to the traditional error-compensation framework. We notice that \cite{xie2020cser} proposes Partial Synchronization which is similar to our local immediate error-compensation framework. However, LIEC-SGD does not need to accumulate the local error and perform Partial Synchronization on it. More importantly, LIEC-SGD enjoys lower communication cost by implementing bidirectional compression.

\subsection{Theoretical Analysis}
We show the theoretical results of our proposed LIEC-SGD for non-convex problems based on the assumptions listed before. We would analyze three cases according to adopting Assumptions \ref{assu3}, \ref{assu4} and \ref{assu5} or not. We need to point out that not adopting Assumption \ref{assu3} is challenging since most previous works use it \cite{stich2018sparsified,tang2019doublesqueeze,zheng2019communication,xie2020cser,xu2021step}. Due to limited space, we include the proof details in the supplementary material.

$\bullet$ \textbf{Case 1: analysis with Assumptions~\ref{assu1}-\ref{assu2}}
\begin{theorem}\label{theo_1}
Consider $f(x)$ under Assumptions~\ref{assu1}-\ref{assu2}. If the learning rate $\eta < \frac{\delta}{10L}$ , then we have the following result for Algorithm~\ref{LIEC}
\begin{eqnarray}
    &&\frac{1-8\eta L}{T} \sum_{t=0}^{T-1}\mathbb{E}\|\nabla f(\Bar{x}_t)\|^2 \nonumber \\
    &\leq&  \frac{8(f(x_0) - f^*)}{\eta T} + (\frac{80\eta^2 L^2}{\delta^2}+\frac{4\eta L}{N}) \sigma^2 \nonumber
\end{eqnarray}
\end{theorem}

\begin{remark}
In the obtained upper bound, $\frac{8(f(x_0) - f^*)}{\eta T}$ depends on the initial value of objective function. $\frac{4\eta L \sigma^2}{N}$ is incurred by the randomness in stochastic gradient. $\frac{80\eta^2 L^2 \sigma^2}{\delta^2}$ represents the bias caused by our Local Immediate Error Compensation with parameter $\delta$, mainly consists of the drift between average and local models and the error saved on the server.
\end{remark}

\begin{corollary}\label{coro_1}
Consider $f(x)$ under Assumptions~\ref{assu1}-\ref{assu2}. If we set the learning rate as
\begin{equation}
    \eta = \frac{1}{\sqrt{\frac{T}{N}}+L+\frac{T^{1/3}}{\delta^{2/3}}}, \nonumber
\end{equation}
then for all $T>0$, we have the following convergence rate for Algorithm~\ref{LIEC}
\begin{equation}
    \frac{1}{T}\sum_{t=0}^{T-1}\mathbb{E} \|\nabla f(\Bar{x}_t)\|^2 \leq O\bigg(\frac{1}{\sqrt{NT}}+\frac{1}{\delta^{2/3} T^{2/3}}+\frac{1}{T}\bigg). \nonumber
\end{equation}
\end{corollary}

\begin{remark}\label{remark2}
We compare the convergence rates of LIEC-SGD and previous works:
\begin{itemize}
    \item MEM-SGD\footnote{This result can be obtained by following the proof in~\cite{zheng2019communication}.}:
    \begin{equation}
        \frac{1}{T}\sum_{t=0}^{T-1}\mathbb{E} \|\nabla f(x_t)\|^2 \leq O\bigg(\frac{1}{\sqrt{NT}}+\frac{1}{\delta^{2/3} T^{2/3}}+\frac{1}{T}\bigg), \nonumber
    \end{equation}
    \item DoubleSqueeze:
    \begin{equation}
        \frac{1}{T}\sum_{t=0}^{T-1}\mathbb{E} \|\nabla f(x_t)\|^2 \leq O\bigg(\frac{1}{\sqrt{NT}}+\frac{1}{\delta^{4/3} T^{2/3}}+\frac{1}{T}\bigg). \nonumber
    \end{equation}
\end{itemize}

Compared to DoubleSqueeze, our algorithm reduces the term corresponding to the error caused by the compression from $O(\frac{1}{\delta^{4/3} T^{2/3}})$ to $O(\frac{1}{\delta^{2/3} T^{2/3}})$. On the other hand, LIEC-SGD involves consistent convergence rate with that of MEM-SGD, but enjoys lower communication cost by bidirectional compression. Thus, our proposed LIEC-SGD inherits the advantages from both unidirectional and bidirectional compression. In addition, we do not need to assume gradient bounded (Assumption \ref{assu3}) to obtain the result while the other two algorithms need.

Simultaneously, our result is tighter than \cite{philippenko2021preserved} (Theorem S11) since $O(1/\sqrt{NT} + 1/\delta^{2/3}T^{2/3})$ is tighter than $O(\sqrt{(1+\omega)/NT})$ when T is sufficiently large. The dominant term of our convergence rate is equal to that of \cite{gruntkowska2023ef21} (Theorem 6.2). Compared to the upper bound of \cite{huang2022lower}, when $T < O(\frac{1}{\delta})$, the dominant term $O(\frac{1}{\delta^{2/3} T^{2/3}})$ of LIEC-SGD is tighter than that $O(\frac{1}{\delta T}$) of \cite{huang2022lower}. When $O(\frac{1}{\delta}) < T < O(\frac{N^3}{\delta^4})$, the dominant term of Neolithic is tighter than that of LIEC-SGD. When $T > O(\frac{N^3} {\delta^4})$, the dominant terms of them two are same as $O(\frac{1}{\sqrt{NT}})$. However, the recursive compression of \cite{huang2022lower} brings extra computation and communication cost, which increases the total time cost. We would reveal this point in experiments.
\end{remark}

\begin{remark}\label{remark3}
This result indicates that when $T \geq O(\frac{N^3}{\delta^4})$, the linear speedup is achieved. As a comparison, DoubleSqueeze needs $T \geq O(\frac{N^3}{\delta^8})$ iterations to achieve the linear speedup. 
\end{remark}

Here we consider the effect of $\delta$ when calculating the threshold of $T$ after which the linear speedup is guaranteed, same as the presentation of results in \cite{huang2022lower}. We note that $\delta$ could represent the estimation or the upper bound of the sparsity of some sparsification operators. Thus, our consideration is reasonable since the sparsity of compressed gradients can not be arbitrarily low, otherwise, the convergence of the algorithm could not be guaranteed if only a few coordinates remain.

$\bullet$ \textbf{Case 2: analysis with Assumptions~\ref{assu1}-\ref{assu4}}

In this case, we provide the analysis for special compressors that satisfy Assumption \ref{assu4}, to show their benefits in theory.
\begin{lemma}\label{lemma1}
If $f(x)$ in Algorithm~\ref{LIEC} satisfies Assumptions~\ref{assu3} and \ref{assu4}, then we have that
\begin{equation}
    \mathbb{E}\|e_t\|^2 \leq \frac{4(1-\delta)(2(2-\delta)+\delta^2 (N-1))M^2}{\delta^2 N}. \nonumber
\end{equation}
\end{lemma}

\begin{remark}\label{remark4}
As a comparison, the corresponding results for the error variables in algorithms with traditional error-compensation 
strategies are as follows~\cite{zheng2019communication}
\begin{itemize}
    \item MEM-SGD:
    \begin{equation}
        \mathbb{E}\|\frac{1}{N}\sum_{i=1}^N e_t^i\|^2 \leq \frac{4(1-\delta)M^2}{\delta^2}, \nonumber
    \end{equation}
    \item DoubleSqueeze:
    \begin{equation}
        \mathbb{E}\|\frac{1}{N}\sum_{i=1}^N e_t^i + e_t\|^2 \leq \frac{8(1-\delta)M^2}{\delta^2}[1+\frac{16}{\delta^2}]. \nonumber
    \end{equation}
\end{itemize}
Focusing on the parameter $\delta$, this comparison indicates that under similar assumptions (although LIEC-SGD additionally adopts Assumption \ref{assu4}, it only influences the constant coefficient), LIEC-SGD controls the bound of the norm of the error variable in the same order as traditional error-compensation algorithms with unidirectional compression, much less than that with bidirectional compression. 
\end{remark}

\begin{lemma}\label{lemma2}
If $f(x)$ in Algorithm~\ref{LIEC} satisfies Assumptions~\ref{assu3} and \ref{assu4}, then we have that
\begin{equation}
    \frac{1}{N}\sum_{i=1}^N \mathbb{E}\|\Bar{x}_t-x_t^i\|^2 \leq \frac{(1-\delta)(1-\delta+\delta^2)}{\delta^2}\eta^2 M^2. \nonumber
\end{equation}
\end{lemma}
This result indicates that although LIEC-SGD yields differences among the local models, the bound of its norm is in the same order as that of the error variable $e_t$, if ignoring the learning rate $\eta$.

\begin{theorem}\label{theo_2}
Consider $f(x)$ under Assumptions~\ref{assu1}-\ref{assu4}. If the learning rate $\eta < \frac{1}{2L}$ , then we have the following result for Algorithm~\ref{LIEC}
\begin{eqnarray}
&&\frac{1-2\eta L}{T}\sum_{t=0}^{T-1} \mathbb{E} \|\nabla f(\Bar{x}_t)\|^2 \nonumber \\
&\leq& \frac{2(f(x_0)-f^*)}{\eta T} + \frac{2(1-\delta)\mathcal{D}\eta^2 L^2 M^2}{\delta^2} + \frac{\eta L \sigma^2}{N}, \nonumber
\end{eqnarray}
where $\mathcal{D} = 1-\delta+\delta^2+\frac{2(4+\delta^2(N-1))}{N}$.
\end{theorem}

Setting the learning rate $\eta$ as that in Corollary~\ref{coro_1} would also yield the convergence rate $O(\frac{1}{\sqrt{NT}}+\frac{1}{\delta^{2/3} T^{2/3}}+\frac{1}{T})$, we do not repeat this conclusion here. 

\begin{remark}\label{remark5}
Compared to Theorem \ref{theo_1}, the constant coefficient of the bound in Theorem \ref{theo_2} is reduced benefited from Assumption \ref{assu4}. In addition, Lemma \ref{lemma1} acts as an intermediate result to indicate that controlling the norm of error effectively is the core reason for the advantage of LIEC-SGD over previous works under the same assumptions in theory.
\end{remark}

$\bullet$ \textbf{Case 3: analysis with Assumption~\ref{assu1}, \ref{assu2} \& \ref{assu5}}
\begin{theorem}\label{theo3}
    Consider $f(x)$ under Assumptions \ref{assu1}, \ref{assu2} and \ref{assu5}. By setting the learning rate the same as Corollary \ref{coro_1}, i.e.
    \begin{equation}
        \eta = \frac{1}{\sqrt{\frac{T}{N}}+L+\frac{T^{1/3}}{\delta^{2/3}}}, \nonumber
    \end{equation}
    then we have the following convergence rate for Algorithm~\ref{LIEC}
    \begin{equation}
    \frac{1}{T}\sum_{t=0}^{T-1}\mathbb{E} \|\nabla f(\Bar{x}_t)\|^2 \leq O\bigg(\frac{1}{\sqrt{NT}}+\frac{1}{\delta^{2/3} T^{2/3}}+\frac{1}{T}\bigg). \nonumber
    \end{equation}
\end{theorem}

\section{Experiments}

\begin{table}[!b]
\caption{Averaged value of $\delta$ of running LIEC-SGD in experiments.}\label{delta}
    \centering
    \begin{tabular}{c|cccc}
    \toprule
     & \makecell[c]{CIFAR-10\\SignSGD} & \makecell[c]{CIFAR-10\\Blockwise} & \makecell[c]{CIFAR-100\\SignSGD} & \makecell[c]{CIFAR-100\\Blockwise} \\
     \midrule
     Worker & $0.24\pm0.01$ & $0.33\pm0.02$ & $0.31\pm0.01$ & $0.35\pm0.02$ \\
     Server & $0.40\pm0.01$ & $0.40\pm0.01$ & $0.39\pm0.01$ & $0.40\pm0.01$ \\
     \bottomrule
    \end{tabular}
\end{table}

\begin{figure*}[!t]
\centering
\includegraphics[scale=0.29]{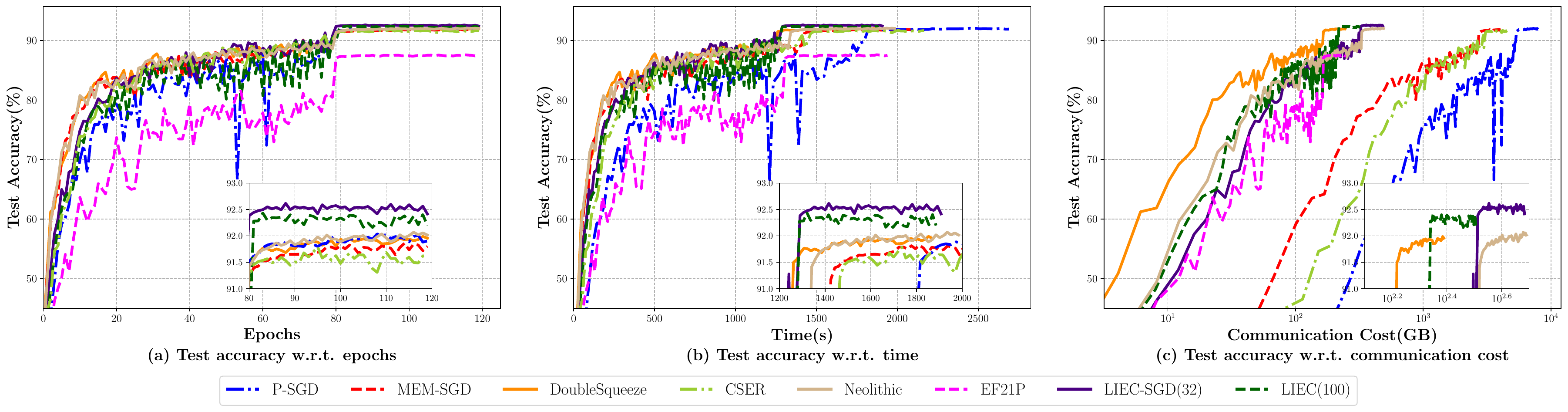}
\vspace{-0.2cm}
\caption{Training ResNet18 on CIFAR-10 with operator SignSGD. (a): Test accuracy w.r.t. epochs. (b): Test accuracy w.r.t. wall-clock time. (c): Test accuracy w.r.t. communication cost.}
\label{fig1}
\end{figure*}

\begin{figure*}[!t]
\centering
\includegraphics[scale=0.29]{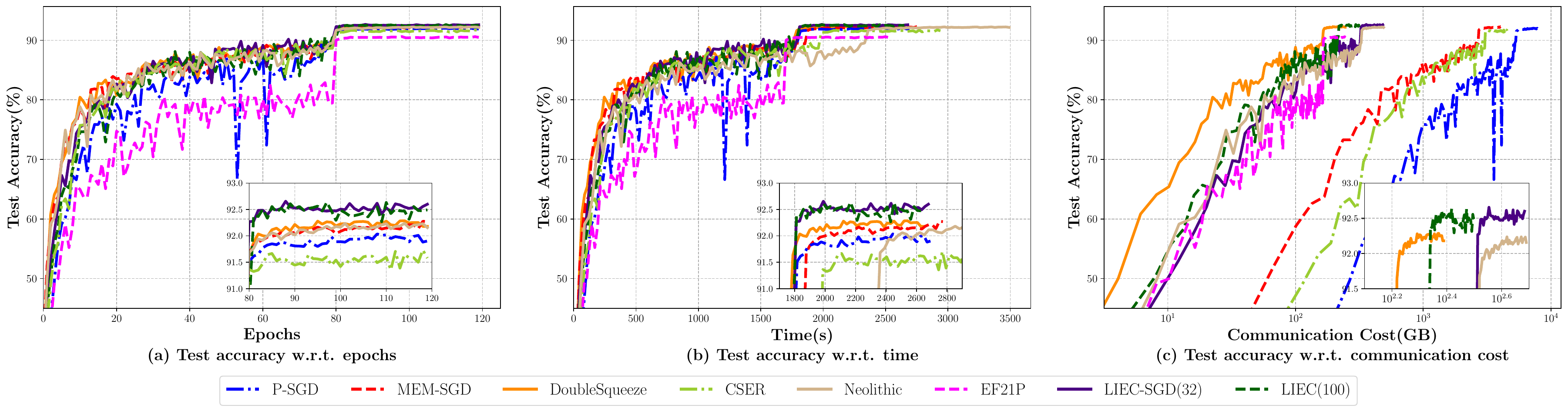}
\vspace{-0.2cm}
\caption{Training ResNet18 on CIFAR-10 with operator Blockwise-SignSGD. (a): Test accuracy w.r.t. epochs. (b): Test accuracy w.r.t. wall-clock time. (c): Test accuracy w.r.t. communication cost.}
\label{fig2}
\end{figure*}

\begin{table*}[!t]
  \vspace{-0.2cm}
  \caption{Communication Cost and Test Accuracy of All Algorithms in Different Experimental Settings.}
  \label{exp}
  \centering
  \begin{threeparttable}
  \begin{tabular}{ccccccc}
    \toprule
    \multirow{3}{*}{Algorithm} & \multicolumn{3}{c}{CIFAR-10} & \multicolumn{3}{c}{CIFAR-100} \\
    \cmidrule(r){2-4} \cmidrule(r){5-7}
    & \multirow{2}{*}{Comm. Cost\tnote{a}} & \multicolumn{2}{c}{Acc(\%)\tnote{b}} & \multirow{2}{*}{Comm. Cost} & \multicolumn{2}{c}{Acc(\%)} \\
    \cmidrule(r){3-4} \cmidrule(r){6-7}
    & & SignSGD & Blockwise-SignSGD & & SignSGD & Blockwise-SignSGD\\
    \midrule
    P-SGD & 681.83 & 92.11$\pm$0.17 & 92.11$\pm$0.17 & 1298.79 & 71.34$\pm$0.13 & 71.34$\pm$0.13   \\
    MEM-SGD & 351.57 & 91.94$\pm$0.16 & 92.34$\pm$0.12 & 669.69 & 70.48$\pm$0.10  & 71.20$\pm$0.31 \\
    DoubleSqueeze & 21.31 & 92.04$\pm$0.09 & 92.37$\pm$0.04 & 40.59 & 70.82$\pm$0.23 & 71.17$\pm$0.34 \\
    CSER & 392.52 & 91.87$\pm$0.07 & 91.92$\pm$0.14 & 753.40 & 70.85$\pm$0.10 & 70.73$\pm$0.38 \\
    Neolithic & 42.62 & 92.12$\pm$0.07 & 92.30$\pm$0.07 & 81.18 & 70.86$\pm$0.45 & 71.33$\pm$0.26 \\
    EF21P & 21.31 & 87.67$\pm$0.25 & 90.67$\pm$0.14 & 40.59 & 62.19$\pm$0.31 & 69.23$\pm$0.34 \\
    LIEC-SGD(32) & 41.95 & \textbf{92.65$\pm$0.14} & \textbf{92.68$\pm$0.04} & 79.91 & \textbf{72.43$\pm$0.41} & \textbf{72.78$\pm$0.15}\\
    LIEC-SGD(100) & 27.91 & \textbf{92.52$\pm$0.11} & \textbf{92.70$\pm$0.09} & 53.16 & \textbf{72.73$\pm$0.22} & \textbf{72.77$\pm$0.18}\\
    \bottomrule
  \end{tabular}
  \begin{tablenotes}
    \item[a] Average communication cost (MB) per iteration.
    \item[b] Averaged best test accuracy.
    \end{tablenotes}
    \end{threeparttable}
    \vspace{-0.2cm}
\end{table*}

We validate our theoretical results by comparing our algorithm with several baselines through training deep neural networks. We train ResNet18~\cite{he2016deep} on the CIFAR-10 dataset~\cite{krizhevsky2009learning}, ResNet34 on the CIFAR-100 dataset and ResNet50 on the Tiny ImageNet dataset respectively for the image classification task. All experiments are repeated over 3 random seeds. We organize $N=8$ workers in total. The experiments are conducted on a machine with NVIDIA 3090 GPUs. Our code is available at https://github.com/chengyif/LIEC-SGD.

\subsection{Baselines and Hyper-parameters on CIFAR dataset}
We compare LIEC-SGD with the following baselines: P-SGD, MEM-SGD\cite{stich2018sparsified}, DoubleSqueeze\cite{tang2019doublesqueeze}, CSER\cite{xie2020cser}, Neolithic\cite{huang2022lower}, EF21P (+DCGD, as Algorithm 2 in \cite{gruntkowska2023ef21}). SignSGD and Blockwise-SignSGD are adopted as the compression operator. When using Blockwise-SignSGD, we split the gradient vectors into 10 blocks for ResNet18 and 18 blocks for ResNet34. The model averaging period $1/\lfloor \frac{1}{\delta} \rfloor$ in LIEC-SGD depends on the value of $\delta$. However, the value of $\delta$ varies throughout the training, and the averaged value also changes depending on the dataset, compression operator, operation side (worker or server) and algorithm (we calculate the exact values of $\delta$ of running LIEC-SGD and summarize them in Table \ref{delta}). For these reasons, we consider two options 32 and 100 as average period to test the performance. The period of error reset in CSER is set to 8. We follow \cite{huang2022lower} to set the compression rounds of Neolithic to 2. The batch size on each worker is 128. Same to \cite{tang2019doublesqueeze}, we set the momentum parameter to zero to compare the performance more directly. The learning rate is set to 0.1 initially as suggested in \cite{he2016deep} and reduced by a factor of 10 at the $80$-th epoch. The weight decay parameter is 0.0005. We run each algorithm for 120 epochs.

\subsection{Results of training ResNet18 on CIFAR-10}
We first list the average communication cost in each iteration of all the baselines in Table~\ref{exp} (the communication costs of SignSGD and Blockwise-SignSGD are the same when only keeping two decimal places). The number in the parenthesises represents the model averaging period of LIEC-SGD. Obviously, the algorithms that perform bidirectional compression (DoubleSqueeze, Neolithic, EF21P and LIEC-SGD) have a huge advantage in this indicator. 

\begin{figure*}[!t]
\centering
\includegraphics[scale=0.29]{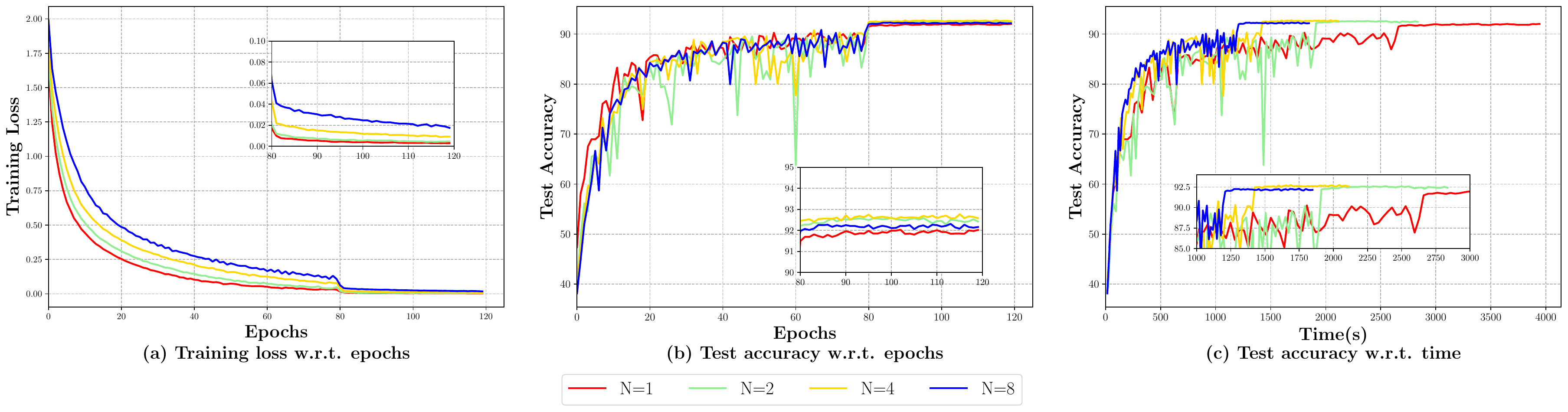}
\vspace{-0.2cm}
\caption{Using LIEC-SGD to train ResNet18 on CIFAR-10 with different number of workers. (a): Training loss w.r.t. epochs. (b): Test accuracy w.r.t. epochs. (c): Test accuracy w.r.t. wall-clock time.}
\label{speedup}
\vspace{-0.4cm}
\end{figure*}

\begin{figure}[!b]
\centering
\includegraphics[scale=0.12]{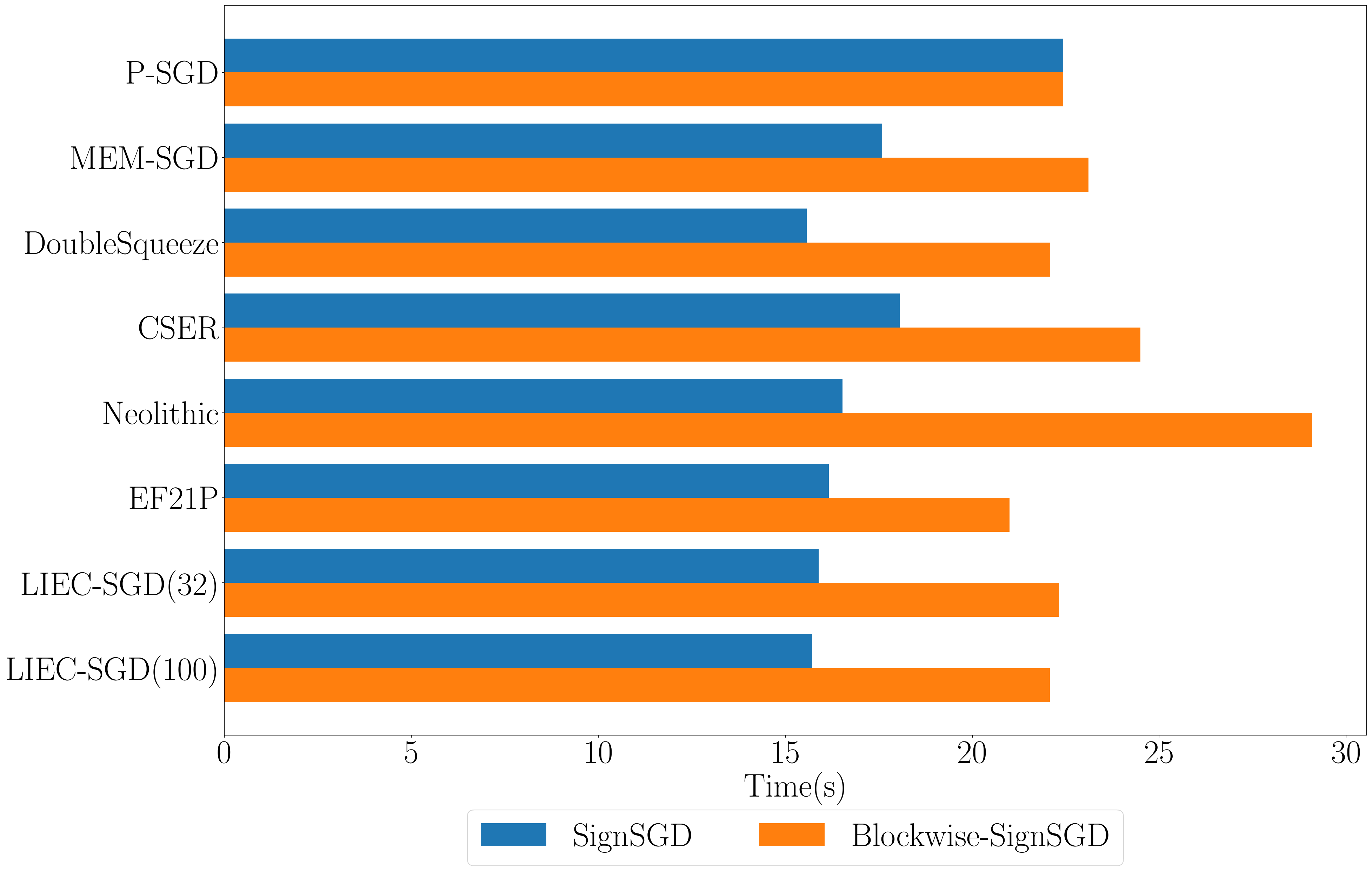}
\caption{Average time cost per epoch when training CIFAR-10.}
\label{time_epoch}
\end{figure}

We show the detailed results of using compressors SignSGD and Blockwise-SignSGD in Figs.~\ref{fig1} and \ref{fig2} respectively. Subfigures (a) plot the test accuracy with respect to (w.r.t.) epochs. The curve of LIEC-SGD increases a little more slowly than MEM-SGD and DoubleSqueeze in the first 30 epochs, however, it catches up in the middle of training. After the 80-th epoch, LIEC-SGD is slightly ahead of other methods. The best test accuracies of all algorithms are summarized in Table~\ref{exp}. LIEC-SGD achieves the highest two test accuracies whatever the compressor is used, thus our algorithm is robust in the choices of compressors. EF21P has an obvious gap to other baselines. We think this is due to the excellent theoretical result of EF21P \cite{gruntkowska2023ef21} relies on that the compressor $\mathbb{U}(\omega)$ is unbiased, which is not satisfied by our adopted compressor. This indicates that our approach is applicable to more compressors. The comparison of the accuracy curves w.r.t. wall-clock time is shown in subfigures (b). DoubleSqueeze converges fastest since it adopts bidirectional compression throughout the training. LIEC-SGD(100) converges a little more early than LIEC-SGD(32) as the former involves fewer rounds of model averaging. These three methods converge at a similar rate and spend much less time than unidirectional compression methods and P-SGD. Though Neolithic also adopts bidirectional compression, it involves multiple rounds of compression and communication in one iteration which bring extra time cost. Consequently, Neolithic spends the most time when using Blockwise-SignSGD compressor, which is not efficient in the time cost term. We calculate the average time cost of each epoch for all baselines and plot the result in Fig.~\ref{time_epoch}. Subfigures (b) best represent the practical performance of all baselines. We show the comparison of test accuracy w.r.t. communication cost in subfigures (c). The curves reveal the advantage of bidirectional compression methods. We mainly report the test accuracy here, and the training and test loss could be referred to the supplementary material.

\subsection{Validation of linear speedup}
According to Remark~\ref{remark3}, LIEC-SGD achieves the linear speedup when $T$ is sufficiently large. In this part, we validate this property by adopting LIEC-SGD to train Resnet18 on CIFAR-10 with different number of workers $N \in \{1,2,4,8\}$. We choose SignSGD as the compression operator and set the model averaging period to 100. Since the number of iterations $T$ is inversely proportional to the number of workers $N$ when the algorithms are run with a fixed number of epochs, the learning rate suggested by Corollary~\ref{coro_1} is approximately proportional to $N$. We set the learning rate to $N/80$. 

From subfigures (a) and (b) in the Fig.~\ref{speedup}, we can find that the training loss and test accuracy curves of different number of workers all converge to the same level. This result verifies the linear speedup of LIEC-SGD since each worker only involves $1/N$ number of total epochs. Subfigure (c) which plots the test accuracy w.r.t. wall-clock time more clearly shows the speedup effect w.r.t. number of workers of our proposed algorithm. 

\begin{table}[!b]
  \vspace{-0.2cm}
  \caption{Training Time Speedup Ratio of All Algorithms.}
  \label{ratio}
  \centering
  \begin{tabular}{ccccc}
    \toprule
    \multirow{2}{*}{Algorithm} & \multicolumn{2}{c}{CIFAR-10} & \multicolumn{2}{c}{CIFAR-100} \\
    \cmidrule(r){2-3} \cmidrule(r){4-5}
    & SignSGD & Blockwise & SignSGD & Blockwise\\
    \midrule
    P-SGD & $1\times$ & $1\times$ & $1\times$ & $1\times$ \\
    MEM-SGD & $1.275\times$ & $0.971\times$ & $1.454\times$ & $0.988\times$ \\
    DoubleSqueeze & $1.441\times$ & $1.016\times$ & $1.737\times$ & $1.064\times$ \\
    CSER & $1.242\times$ & $0.915\times$ & $1.376\times$ & $0.926\times$ \\
    Neolithic & $1.357\times$ & $0.771\times$ & $1.589\times$ & $0.749\times$ \\
    EF21P & $1.388\times$ & $1.068\times$ & $1.637\times$ & $1.126\times$ \\
    LIEC-SGD(32) & $1.399\times$ & $1.005\times$ & $1.676\times$ & $1.044\times$ \\
    LIEC-SGD(100) & $1.428\times$ & $1.016\times$ & $1.721\times$ & $1.057\times$ \\
    \bottomrule
  \end{tabular}
\end{table}

\begin{figure*}[!t]
\centering
\includegraphics[scale=0.29]{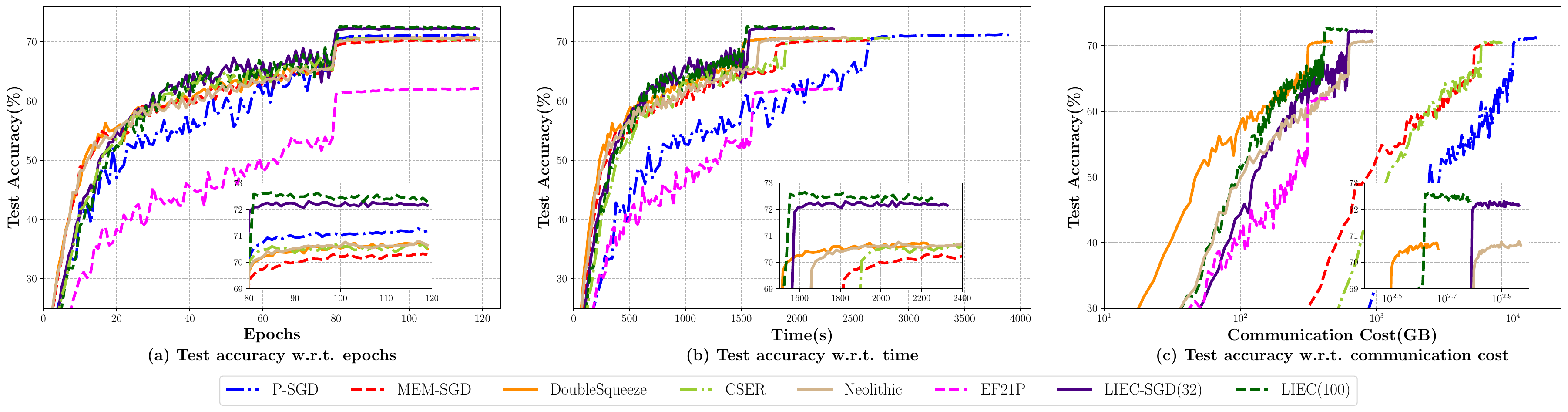}
\vspace{-0.2cm}
\caption{Training ResNet34 on CIFAR-100 with operator SignSGD. (a): Test accuracy w.r.t. epochs. (b): Test accuracy w.r.t. wall-clock time. (c): Test accuracy w.r.t. communication cost.}
\label{fig3}
\end{figure*}

\begin{figure*}[!t]
\centering
\includegraphics[scale=0.29]{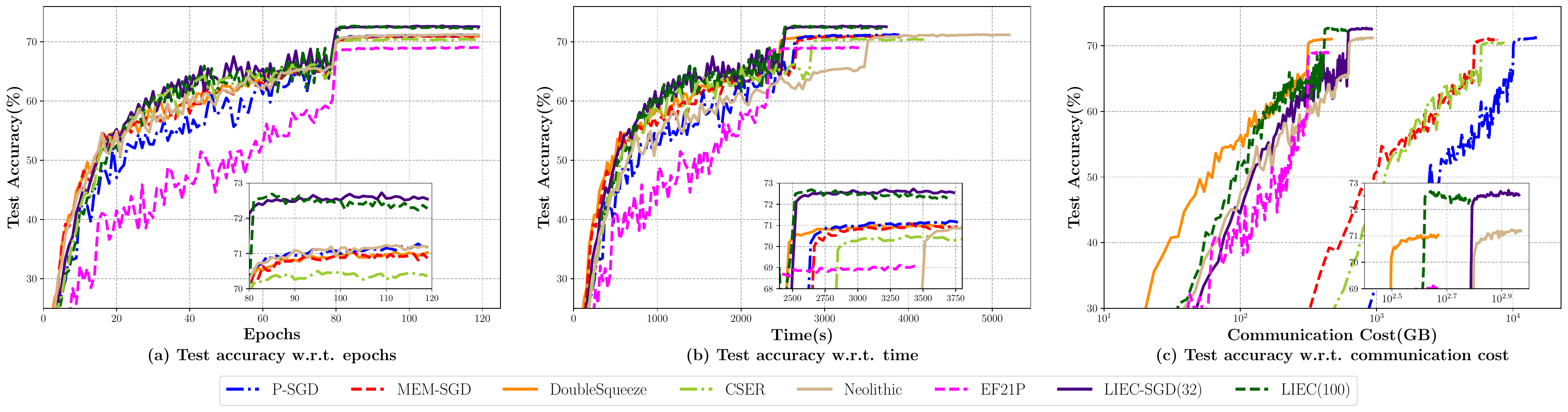}
\vspace{-0.2cm}
\caption{Training ResNet34 on CIFAR-100 with operator Blockwise-SignSGD. (a): Test accuracy w.r.t. epochs. (b): Test accuracy w.r.t. wall-clock time. (c): Test accuracy w.r.t. communication cost.}
\label{fig4}
\vspace{-0.2cm}
\end{figure*}

\subsection{Results of training ResNet34 on CIFAR-100}

The experimental comparison is shown in Figs.~\ref{fig3} and \ref{fig4}. From the results in subfigures (a), we observe that after the 40-th epoch, LIEC-SGD converges faster than all other methods. This result shows the high efficiency of LIEC-SGD, which matches our theoretical conclusions and reveals its significant generalization ability again. We summarize the best test accuracy of all algorithms in Table~\ref{exp} and LIEC-SGD has greater advantages in this experiment. We note that several algorithms outperform P-SGD which transmits full precision gradient. This phenomenon also appears in the experiments of some related works~\cite{tang2019doublesqueeze,xu2021step}. This may be because that inaccurate gradient could improve the generalization ability~\cite{neelakantan2015adding,orvieto2022anticorrelated}.

In the subfigures (b) which plot the text accuracy w.r.t. time, LIEC-SGD is still time-saving while achieving the best testing performance. We summarize the speedup ratios of the total training time of all algorithms compared to P-SGD in Table~\ref{ratio}. The speedup ratios of DoubleSqueeze, EF21P and LIEC-SGD, which all conduct bidirectional compression, are larger than 1 in all cases. On the contrary, the algorithms with unidirectional compression (MEM-SGD and CSER) may face the phenomenon which is mentioned in Section 2.2 that the extra computation time cost brought by compression may exceed the communication time reduction. To explain the cause of this result, we make a more detailed analysis on the time cost. We find that the server needs to implement uncompression on the compressed gradients serially while workers could uncompress the compressed vectors sent by the server parallelly which takes very little time. As a consequence, although in the direction from workers to server, the extra computation overhead exceeds the communication reduction, the huge advantage of communication reduction over the extra computation in the direction from server to workers could outweigh this negative factor. Overall, bidirectional compression is necessary for efficient distributed learning. Finally, subfigures (c) show that LIEC-SGD and DoubleSqueeze achieve great test accuracy while spending a small amount of communication cost.

\subsection{Effects of Immediate Error-Compensation}
\begin{figure*}[!t]
\centering
\includegraphics[scale=0.29]{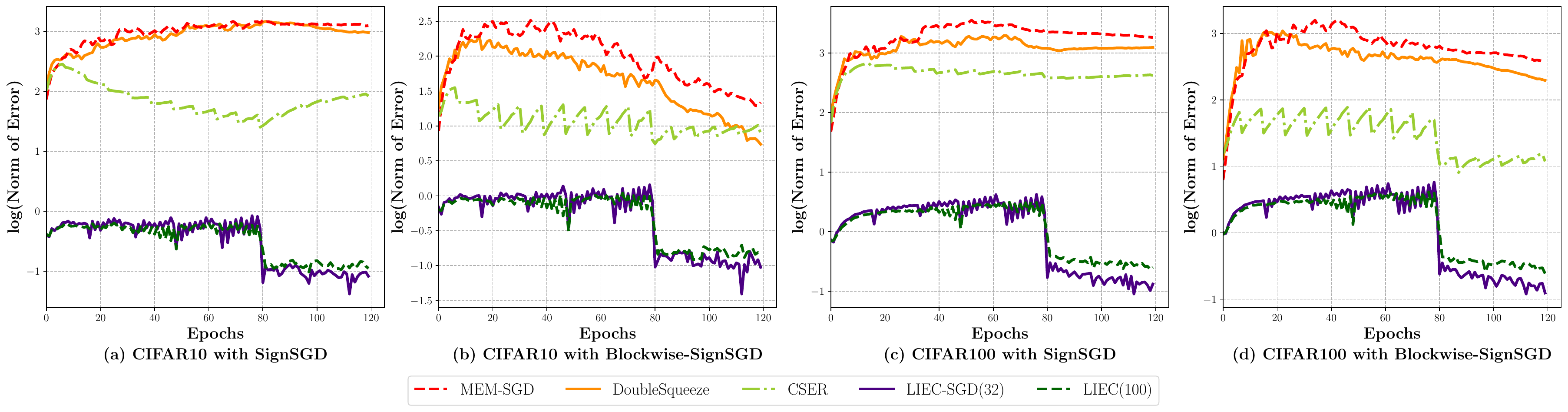}
\vspace{-0.2cm}
\caption{Logarithm of the norm of error variables in different experiments.}
\label{error}
\end{figure*}

In this subsection, we show the logarithm of the $l2$-norm of error variables (as defined in Lemma~\ref{lemma1} and Remark~\ref{remark4}) w.r.t. epochs in Fig. \ref{error}. The comparison indicates that LIEC-SGD observably reduces the norm of the error variables which alleviates the "remained gradients" dilemma mentioned before greatly. This conclusion is in line with our original intention to design the local immediate error-compensation framework and the conclusion in Remark \ref{remark5} that controlling the norm of error variable is the core factor for the fast convergence rate.

\begin{figure*}[!t]
\centering
\includegraphics[scale=0.29]{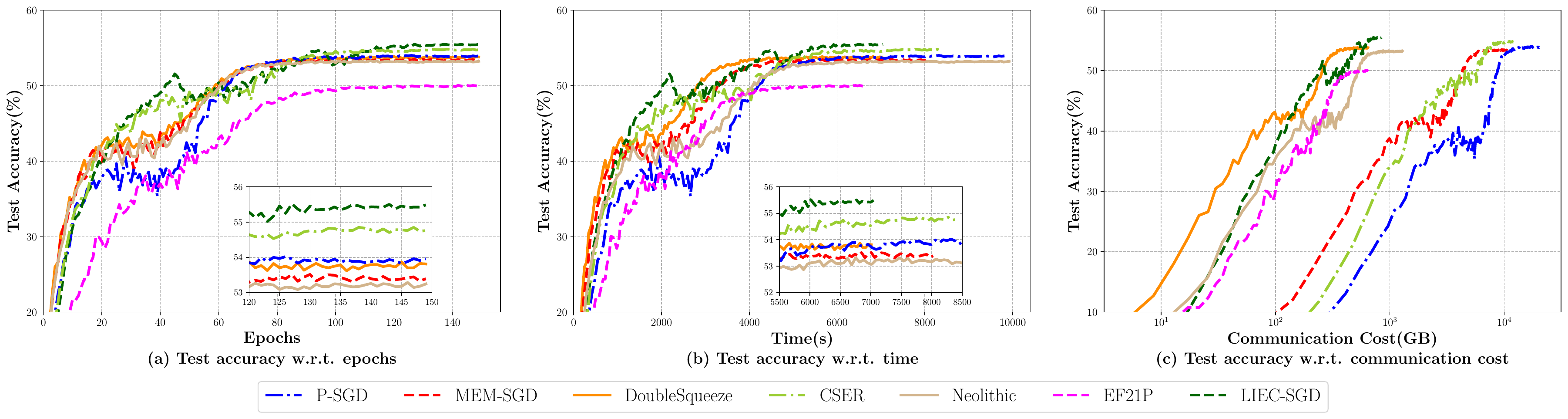}
\vspace{-0.2cm}
\caption{Training ResNet50 on Tiny ImageNet with operator Blockwise-SignSGD. (a): Test accuracy w.r.t. epochs. (b): Test accuracy w.r.t. wall-clock time. (c): Test accuracy w.r.t. communication cost.}
\label{fig5}
\vspace{-0.2cm}
\end{figure*}

\begin{table}[!t]
  \vspace{-0.2cm}
  \caption{Best test accuracy (\%) of all algorithms on Tiny ImageNet.}
  \label{acc_tiny}
  \centering
  \begin{tabular}{c|c|c}
    \toprule
    Algorithm & \makecell[c]{Averaged Best\\Test Accuracy} & \makecell[c]{Training Time\\Speedup Ratio} \\
    \midrule
    P-SGD & 54.14$\pm$0.14 & $1\times$ \\
    MEM-SGD & 53.70$\pm$0.29 & $1.223\times$ \\
    DoubleSqueeze & 53.96$\pm$0.34 & $1.400\times$\\
    CSER & 54.97$\pm$0.09 & $1.171\times$\\
    Neolithic & 53.54$\pm$0.34 & $0.989\times$\\
    EF21P & 50.15$\pm$0.50 & $1.487\times$\\
    LIEC & \textbf{55.61$\pm$0.20} & $1.393\times$\\
    \bottomrule
  \end{tabular}
  \vspace{-0.2cm}
\end{table}

\subsection{Hyper-parameters on Tiny ImageNet dataset}
In this part, we compare the performances of the algorithms same to the experiments on CIFAR dataset. We adopt Blockwise-SignSGD as the compression operator. The gradient vectors are divided into 14 blocks. The model averaging period in LIEC-SGD and the period of error reset in CSER are set to 100 and 16, respectively. The batch size per worker is 128. We set the initial learning rate to 0.2 and reduce it to the minimal learning rate 0.001 with the cosine learning rate scheduler. All algorithms are run for 150 epochs.

\subsection{Results of training ResNet50 on Tiny ImageNet}
We plot the results of training Tiny ImageNet dataset in Figure~\ref{fig5}. In the subfigure (a), we show the test accuracy curves w.r.t. the epochs. We can find the in the 40-50th epochs, LIEC-SGD shows a distinct advantage over other baselines. At the end of the training, the curve of LIEC-SGD converges to the highest level among all methods. CSER achieves the second highest accuracy benefited from the frequent error average and reset. We then plot the test accuracy w.r.t. wall-clock time in the subfigure (b). The methods LIEC-SGD, EF21P and DoubleSqueeze, which take the bidirectional compression still spend the least training time, specifically about $70\%$ of P-SGD, thus showing the advantage of saving time cost. The unidirectional methods and Neolithic obviously take more wall-clock time in the learning process. Detailed results are listed in Table~\ref{acc_tiny}.

\section{Conclusion}
In this paper, we propose the LIEC-SGD algorithm for large-scale distributed learning. LIEC-SGD adopts the local immediate error-compensation framework which compensates the local compression error on the worker to the gradient sent from the server to update the model without any delay. For saving the communication cost efficiently, bidirectional compression is performed in our algorithm. We theoretically prove the convergence rate of LIEC-SGD with $\delta$-contraction operator and analyze its advantage over the previous works which adopt the traditional error-compensation framework. Finally, in the experiments on training deep neural networks for the image classification task (including different models and datasets), LIEC-SGD achieves the best test accuracy in all settings while saving much training time, which demonstrates the effectiveness of our proposed algorithm.


%

\appendices


\ifCLASSOPTIONcaptionsoff
  \newpage
\fi



\bibliographystyle{IEEEtran}
\bibliography{references}

\clearpage
In the supplemental material, we present the proof details of the theoretical results and more experimental results.
\section{Theoretical Analysis}
\subsection{Useful Inequalities}
We first introduce some inequalities used in our analysis.
\begin{itemize}
    \item \textbf{Cauchy's Inequality (special case): } $\forall x_i \in R^d$
\begin{equation}
    \|\sum_{i=1}^n x_i\|^2 \leq n \sum_{i=1}^n\|x_i\|^2. \nonumber
\end{equation}
    \item \textbf{Young's Inequality: } $\forall a,b \in R^d, \beta > 0$
\begin{equation}
    \|a+b\|^2 \leq (1+\beta)\|a\|^2 + (1+\frac{1}{\beta})\|b\|^2. \nonumber
\end{equation}
\end{itemize}

\subsection{Proofs of Theoretical Results}
In this section, we will first give the proof details of Theorem 1, then establish the theoretical analysis for Lemmas 1, 2 and Theorem 2. Now we begin the proof. Define the virtual sequence $\{\hat{x}_t\}$ as follows:
\begin{equation}
    \hat{x}_{t+1} = \hat{x}_t- \frac{1}{N} \sum_{i=1}^N \eta \nabla f_i(x_t^i, \xi_t^i). \nonumber
\end{equation}
Then we have that
\setcounter{lemma}{2}
\begin{lemma}\label{lemma3}
For $\{\hat{x}_t\}$ and $\{\Bar{x}_t\}$ in Algorithm 1
\begin{equation}
    \Bar{x}_t-\hat{x}_t = \eta e_t. \nonumber
\end{equation}
\end{lemma}

\begin{IEEEproof}
According to the update rule, we obtain that
\begin{eqnarray}
&&\Bar{x}_t-\hat{x}_t \nonumber \\
&=& \bigg(x_0-\sum_{\tau=0}^{t-1} \frac{1}{N} \sum_{i=1}^N \eta (p_\tau-p_\tau^i+\nabla f_i(x_\tau^i, \xi_\tau^i))\bigg) \nonumber \\
&&-\bigg(x_0-\sum_{\tau=0}^{t-1}\frac{1}{N} \sum_{i=1}^N \eta \nabla f_i(x_\tau^i, \xi_\tau^i)\bigg) \nonumber \\
&=& \frac{1}{N} \sum_{\tau=0}^{t-1} \sum_{i=1}^N \eta (p_\tau^i-p_\tau) \nonumber \\
&=& \sum_{\tau=0}^{t-1} \eta (v_\tau-e_\tau+e_{\tau+1}-v_\tau) \nonumber \\
&=& \eta e_t \nonumber
\end{eqnarray}
\end{IEEEproof}

\begin{lemma}\label{lemma4}
    If $f(x)$ in Algorithm 1 satisfies Assumptions 1 and 2, then we have that
    \begin{equation}
        \mathbb{E}\|\nabla f_i(x_t^i, \xi_t^i)\|^2 \leq \sigma^2 + 3L^2 \mathbb{E}\|\Bar{x}_t-x_t^i\|^2 + 3\mathbb{E}\|\nabla f(\Bar{x}_t)\|^2. \nonumber
    \end{equation}
\end{lemma}
\begin{IEEEproof}
\begin{eqnarray}
    &&\mathbb{E}\|\nabla f_i(x_t^i, \xi_t^i)\|^2 \nonumber \\
    &=& \mathbb{E}\|\nabla f_i(x_t^i, \xi_t^i) - \nabla f_i(x_t^i) + \nabla f_i(x_t^i)\|^2 \nonumber \\
    &\stackrel{(a)}{\leq}& \sigma^2 + \mathbb{E}\|\nabla f_i(x_t^i) - \nabla f_i(\Bar{x}_t) + \nabla f_i(\Bar{x}_t) - \nabla f(\Bar{x}_t) \nonumber \\
    &&+ \nabla f(\Bar{x}_t)\|^2 \nonumber \\
    &\stackrel{(b)}{\leq}& \sigma^2 + 3\mathbb{E}\|\nabla f_i(x_t^i) - \nabla f_i(\Bar{x}_t)\|^2 \nonumber \\
    &&+ 3\mathbb{E}\|\nabla f_i(\Bar{x}_t) - \nabla f(\Bar{x}_t)\|^2 + 3\mathbb{E}\|\nabla f(\Bar{x}_t)\|^2 \nonumber \\
    &\stackrel{(c)}{\leq}& \sigma^2 + 3L^2 \mathbb{E}\|\Bar{x}_t-x_t^i\|^2 + 3\mathbb{E}\|\nabla f(\Bar{x}_t)\|^2, \nonumber
\end{eqnarray}
where (a) comes from Assumption 2 and (b) comes from the Cauchy's Inequality. (c) holds because of Assumption 1 the fact that the data is split uniformly on the workers, thus we could assume $\mathbb{E}\|\nabla f_i(x)-\nabla f(x)\|^2 = 0$.
\end{IEEEproof}

\begin{figure*}[!t]
\normalsize
\setcounter{MYtempeqncnt}{\value{equation}}
\setcounter{equation}{1}
\vspace{-0.8cm}
\begin{eqnarray}\label{lemma6_1}
&&\sum_{i=1}^N \mathbb{E}\|\Bar{x}_t-x_t^i\|^2 \nonumber \\
&=& \eta^2  \sum_{i=1}^N \mathbb{E}\|\sum_{\tau=t'}^{t-1}\nabla f_i(x_\tau^i,\xi_\tau^i)-\mathcal{C}_\delta(\nabla f_i(x_\tau^i,\xi_\tau^i))-(\frac{1}{N}\sum_{j=1}^N \nabla f_j(x_\tau^j,\xi_\tau^j)-\mathcal{C}_\delta(\nabla f_j(x_\tau^j,\xi_\tau^j)))\|^2 \nonumber \\
&=& \eta^2 \bigg(\sum_{i=1}^N \mathbb{E}\|\sum_{\tau=t'}^{t-1}\nabla f_i(x_\tau^i,\xi_\tau^i)-\mathcal{C}_\delta(\nabla f_i(x_\tau^i,\xi_\tau^i))\|^2-\frac{1}{N}\mathbb{E}\|\sum_{\tau=t'}^{t-1} \sum_{j=1}^N \nabla f_j(x_\tau^j,\xi_\tau^j)-\mathcal{C}_\delta(\nabla f_j(x_\tau^j,\xi_\tau^j))\|^2\bigg) \nonumber \\
&\leq& \eta^2 \sum_{i=1}^N \mathbb{E}\|\sum_{\tau=t'}^{t-1}\nabla f_i(x_\tau^i,\xi_\tau^i)-\mathcal{C}_\delta(\nabla f_i(x_\tau^i,\xi_\tau^i))\|^2 \nonumber \\
&\stackrel{(a)}{\leq}& \eta^2 (t-t')\sum_{i=1}^N \sum_{\tau=t'}^{t-1} \mathbb{E}\|\nabla f_i(x_\tau^i,\xi_\tau^i)-\mathcal{C}_\delta(\nabla f_i(x_\tau^i,\xi_\tau^i))\|^2 \nonumber \\
&\stackrel{(b)}{\leq}& (1-\delta)\eta^2 (t-t') \sum_{i=1}^N \sum_{\tau=t'}^{t-1} \mathbb{E}\|\nabla f_i(x_\tau^i, \xi_\tau^i)\|^2 \nonumber \\
&\stackrel{(c)}{\leq}& (1-\delta)\eta^2 (t-t') \sum_{i=1}^N \sum_{\tau=t'}^{t-1} (\sigma^2 + 3L^2 \mathbb{E}\|x_\tau^i - \Bar{x}_\tau\|^2 + 3\mathbb{E}\|\nabla f(\Bar{x}_\tau)\|^2) \nonumber \\
&=& (1-\delta)\eta^2(t-t')((t-t')N \sigma^2 + 3L^2 \sum_{i=1}^N \sum_{\tau=t'}^{t-1} \mathbb{E}\|x_\tau^i - \Bar{x}_\tau\|^2 + 3 \mathbb{E}\|\nabla f(\Bar{x}_\tau)\|^2)
\end{eqnarray}
\hrulefill
\vspace*{4pt}
\end{figure*}

\begin{lemma}\label{lemma5}
If $f(x)$ in Algorithm 1 satisfies Assumptions 1 and 2, we have that
\begin{eqnarray}
    \mathbb{E}\|\frac{1}{N} \sum_{i=1}^N \nabla f_i(x_t^i, \xi_t^i)\|^2 &\leq& \frac{\sigma^2}{N} + \frac{2L^2}{N}\sum_{i=1}^N\mathbb{E}\|x_t^i-\Bar{x}_t\|^2 \nonumber \\
    &&+2\mathbb{E}\|\nabla f(\Bar{x}_t)\|^2. \nonumber
\end{eqnarray}
\end{lemma}

\begin{IEEEproof}
\begin{eqnarray}\label{lemma5_1}
&&\mathbb{E}\|\frac{1}{N} \sum_{i=1}^N \nabla f_i(x_t^i, \xi_t^i)\|^2 \nonumber \\
&=& \mathbb{E}\|\frac{1}{N} \sum_{i=1}^N \nabla f_i(x_t^i, \xi_t^i)-\nabla f_i(x_t^i)+\nabla f_i(x_t^i) \nonumber \\
&&-\nabla f_i(\Bar{x}_t)+\nabla f_i(\Bar{x}_t)\|^2 \nonumber 
\end{eqnarray}

Since ${\xi_t^{i}}'s$ are independent and $\mathbb{E}[\nabla f_i(x_t^i, \xi_t^i)]=\nabla f_i(x_t^i)$, we can obtain that
\setcounter{equation}{5}
\begin{eqnarray}\label{lemma5_2}
&&\mathbb{E}\|\frac{1}{N} \sum_{i=1}^N \nabla f_i(x_t^i, \xi_t^i)\|^2 \nonumber \\
&=& \mathbb{E}\|\frac{1}{N} \sum_{i=1}^N \nabla f_i(x_t^i, \xi_t^i)-\nabla f_i(x_t^i)\|^2 \nonumber \\
&&+\mathbb{E}\|\frac{1}{N} \sum_{i=1}^N\nabla f_i(x_t^i)-\nabla f_i(\Bar{x}_t)+\nabla f_i(\Bar{x}_t)\|^2 \nonumber
\end{eqnarray}
Further, we have
\setcounter{equation}{6}
\begin{eqnarray}
&&\mathbb{E}\|\frac{1}{N} \sum_{i=1}^N \nabla f_i(x_t^i, \xi_t^i)\|^2 \nonumber \\
&\stackrel{(a)}{\leq}& \frac{\sigma^2}{N} + 2\mathbb{E}(\|\frac{1}{N} \sum_{i=1}^N\nabla f_i(x_t^i)-\nabla f_i(\Bar{x}_t)\|^2 +\|\nabla f(\Bar{x}_t)\|^2) \nonumber \\
&\stackrel{(b)}{\leq}& \frac{\sigma^2}{N} + \frac{2L^2}{N}\sum_{i=1}^N\mathbb{E}\|x_t^i-\Bar{x}_t\|^2+2\mathbb{E}\|\nabla f(\Bar{x}_t)\|^2, \nonumber
\end{eqnarray}
where (a) follows Assumption 2 and the Cauchy's Inequality. (b) follows from the Cauchy's Inequality and Assumption 1.
\end{IEEEproof}

\begin{lemma}\label{lemma6}
    If $f(x)$ in Algorithm 1 satisfies Assumptions 1 and 2, the learning rate satisfies $\eta \leq \frac{\delta}{\sqrt{6}L}$, then we have that
    \begin{eqnarray}
        &&\frac{1}{N}\sum_{t=0}^{T-1} \sum_{i=1}^N \mathbb{E} \|\Bar{x}_t - x_t^i\|^2 \nonumber \\
        &\leq& \frac{2(1-\delta)\eta^2 T \sigma^2}{\delta^2} + \frac{6(1-\delta)\eta^2}{\delta^2}\sum_{t=0}^{T-1} \mathbb{E} \|\nabla f(\Bar{x}_t)\|^2. \nonumber
    \end{eqnarray}
\end{lemma}
\begin{IEEEproof}
We define $t'<t$ as the largest iteration index that is a multiple of $\lfloor \frac{1}{\delta} \rfloor$. Following the definition of $t'$, we have that $\Bar{x}_{t'} = x_{t'}^i$. According to the update rule, we have 
\begin{eqnarray}
&&\Bar{x}_t-x_t^i \nonumber \\
&=& \Bar{x}_{t'} - \eta \frac{1}{N}\sum_{j=1}^N \sum_{\tau=t'}^{t-1} (p_\tau-p_\tau^j+\nabla f_j(x_\tau^j,\xi_\tau^j)) \nonumber \\
&&-(x_{t'}^i-\eta \sum_{\tau=t'}^{t-1}(p_\tau-p_\tau^i+\nabla f_i(x_\tau^i,\xi_\tau^i))) \nonumber 
\end{eqnarray}
Further, we have
\begin{eqnarray}
\Bar{x}_t-x_t^i &=& \eta \sum_{\tau=t'}^{t-1}(\nabla f_i(x_\tau^i,\xi_\tau^i)-\mathcal{C}_\delta(\nabla f_i(x_\tau^i,\xi_\tau^i)) \nonumber \\
&&-(\frac{1}{N}\sum_{j=1}^N \nabla f_j(x_\tau^j,\xi_\tau^j)-\mathcal{C}_\delta(\nabla f_j(x_\tau^j,\xi_\tau^j)))). \nonumber
\end{eqnarray}
So we have (\ref{lemma6_1}) on the top of this page, where (a) follows Cauchy's Inequality, (b) comes from Definition 1 and (c) comes from Lemma \ref{lemma4}. Next, we sum (\ref{lemma6_1}) over $t \in \{0,1,...,T-1\}$ and obtain (\ref{lemma6_2}) on the top of next page, where (d) holds because $t-t' \leq \lfloor \frac{1}{\delta} \rfloor \leq \frac{1}{\delta}$. (e) comes from the fact that $\frac{3(1-\delta)\eta^2 L^2}{\delta^2} \leq \frac{3\eta^2 L^2}{\delta^2} \leq \frac{1}{2}$. Rearranging (\ref{lemma6_2}) yields the result.
\end{IEEEproof}

\begin{figure*}[h]
\normalsize
\setcounter{MYtempeqncnt}{\value{equation}}
\setcounter{equation}{2}
\vspace{-0.8cm}
\begin{eqnarray}\label{lemma6_2}
\sum_{t=0}^{T-1} \sum_{i=1}^N \mathbb{E} \|\Bar{x}_t - x_t^i\|^2 &\leq& (1-\delta)\sum_{t=0}^{T-1}\eta^2(t-t')((t-t')N \sigma^2 + 3L^2 \sum_{i=1}^N \sum_{\tau=t'}^{t-1} \mathbb{E}\|x_\tau^i - \Bar{x}_\tau\|^2 + 3N \mathbb{E}\|\nabla f(\Bar{x}_\tau)\|^2) \nonumber \\
&\stackrel{(d)}{\leq}& \frac{(1-\delta)\eta^2 T N \sigma^2}{\delta^2} + \frac{3(1-\delta)\eta^2 L^2}{\delta^2} \sum_{t=0}^{T-1} \sum_{i=1}^N \mathbb{E} \|\Bar{x}_t - x_t^i\|^2 + \frac{3(1-\delta)N\eta^2}{\delta^2}\sum_{t=0}^{T-1} \mathbb{E} \|\nabla f(\Bar{x}_t)\|^2 \nonumber \\
&\stackrel{(e)}{\leq}& \frac{(1-\delta)\eta^2 T N \sigma^2}{\delta^2} + \frac{1}{2} \sum_{t=0}^{T-1} \sum_{i=1}^N \mathbb{E} \|\Bar{x}_t - x_t^i\|^2 + \frac{3(1-\delta)N\eta^2}{\delta^2}\sum_{t=0}^{T-1} \mathbb{E} \|\nabla f(\Bar{x}_t)\|^2.
\end{eqnarray}
\hrulefill
\vspace*{4pt}
\end{figure*}

\begin{figure*}[h]
\normalsize
\setcounter{MYtempeqncnt}{\value{equation}}
\setcounter{equation}{3}
\vspace{-0.6cm}
\begin{eqnarray}\label{lemma7_1}
    &&\mathbb{E}\|e_t\|^2 \nonumber \\
    &\stackrel{(b)}{\leq}& (1-\delta)(1+\beta) \mathbb{E}\|e_{t-1}\|^2 + (1-\delta)(1+1/\beta) \mathbb{E}\|\frac{1}{N} \sum_{i=1}^N p_{t-1}^i\|^2 \nonumber \\
    &\stackrel{(c)}{\leq}& (1-\delta)(1+\beta) \mathbb{E}\|e_{t-1}\|^2 + \frac{(1-\delta)(1+1/\beta)}{N} \sum_{i=1}^N \mathbb{E}\|\mathcal{C}_\delta (\nabla f_i(x_{t-1}^i,\xi_{t-1}^i))\|^2 \nonumber \\
    &=& (1-\delta)(1+\beta) \mathbb{E}\|e_{t-1}\|^2 +\frac{(1-\delta)(1+1/\beta)}{N} \sum_{i=1}^N \mathbb{E}\|\mathcal{C}_\delta (\nabla f_i(x_{t-1}^i,\xi_{t-1}^i)-\nabla f_i(x_{t-1}^i,\xi_{t-1}^i)+\nabla f_i(x_{t-1}^i,\xi_{t-1}^i))\|^2 \nonumber \\
    &\stackrel{(d)}{\leq}& (1-\delta)(1+\beta)\mathbb{E}\|e_{t-1}\|^2 + \frac{2(1-\delta)(2-\delta)(1+\frac{1}{\beta})}{N}\sum_{i=1}^N \mathbb{E}\|\nabla f_i(x_{t-1}^i, \xi_{t-1}^i)\|^2 \nonumber \\
    &\stackrel{(e)}{\leq}& (1-\delta)(1+\beta)\mathbb{E}\|e_{t-1}\|^2 + 2(1-\delta)(2-\delta)(1+\frac{1}{\beta})(\sigma^2 + \frac{3L^2}{N} \sum_{i=1}^N\mathbb{E}\|x_{t-1}^i - \Bar{x}_{t-1}\|^2 + 3\mathbb{E}\|\nabla f(\Bar{x}_{t-1})\|^2) \nonumber \\
    &=& 2(1-\delta)(2-\delta)(1+\frac{1}{\beta})\sigma^2 \frac{1}{1-(1-\delta)(1+\beta)} \nonumber \\
    &&+ 6(1-\delta)(2-\delta)(1+\frac{1}{\beta}) \sum_{\tau=t'}^{t-1}[(1-\delta)(1+\beta)]^{t-1-\tau} (\frac{L^2}{N}\sum_{i=1}^N\mathbb{E}\|x_\tau^i-\Bar{x}_\tau\|^2+\mathbb{E}\|\nabla f(\Bar{x}_\tau)\|^2) 
\end{eqnarray}
\hrulefill
\vspace*{4pt}
\end{figure*}

\begin{lemma}\label{lemma7}
    If $f(x)$ in Algorithm 1 satisfies Assumptions 1 and 2, then we have that
    \begin{eqnarray}
        \sum_{t=0}^{T-1}\mathbb{E}\|e_t\|^2 &\leq& \frac{8(1-\delta)(2-\delta)T\sigma^2}{\delta^2} \nonumber \\
        &&+ \frac{24(1-\delta)(2-\delta)L^2}{\delta^2 N}\sum_{t=0}^{T-1}\sum_{i=1}^N\mathbb{E}\|x_t^i-\Bar{x}_t\|^2 \nonumber \\
        &&+ \frac{24(1-\delta)(2-\delta)}{\delta^2}\sum_{t=0}^{T-1} \mathbb{E}\|\nabla f(\Bar{x}_t)\|^2. \nonumber
    \end{eqnarray}
\end{lemma}

\begin{IEEEproof}
When $(t+1) \% \lfloor \frac{1}{\delta} \rfloor=0$, we have $\|e_t\|=0$. We focus on the iterations where $(t+1) \% \lfloor \frac{1}{\delta} \rfloor\neq0$.
\begin{equation}
\mathbb{E}\|e_t\|^2 = \mathbb{E}\|v_{t-1}-\mathcal{C}_\delta(v_{t-1})\|^2 \stackrel{(a)}{\leq} (1-\delta) \mathbb{E}\|v_{t-1}\|^2 \nonumber 
\end{equation}
where (a) comes from Definition 1. Further, we have (\ref{lemma7_1}) on next page. (b) comes from Young's Inequality. (c) comes from the Cauchy's Inequality. (d) follows from Cauchy's Inequality and Definition 1. (e) comes from Lemma \ref{lemma4}. Summing up (\ref{lemma7_1}) over $\{0, 1, ..., T-1\}$ yields (\ref{lemma7_2}) on the next page, where (f) holds by setting $\beta = \frac{\delta}{2(1-\delta)}$. 
\end{IEEEproof}

\begin{figure*}[h]
\normalsize
\setcounter{MYtempeqncnt}{\value{equation}}
\setcounter{equation}{4}
\vspace{-0.2cm}
\begin{eqnarray}\label{lemma7_2}
    \sum_{t=0}^{T-1}\mathbb{E}\|e_t\|^2 &\leq&  \frac{2(1-\delta)(2-\delta)(1+\frac{1}{\beta})T\sigma^2}{1-(1-\delta)(1+\beta)} \nonumber \\
    &&+ 6(1-\delta)(2-\delta)(1+\frac{1}{\beta}) \sum_{t=0}^{T-1}\sum_{\tau=t'}^{t-1}[(1-\delta)(1+\beta)]^{t-1-\tau} (\frac{L^2}{N}\sum_{i=1}^N\mathbb{E}\|x_\tau^i-\Bar{x}_\tau\|^2+\mathbb{E}\|\nabla f(\Bar{x}_\tau)\|^2) \nonumber \\
    &\leq& \frac{2(1-\delta)(2-\delta)(1+\frac{1}{\beta})T\sigma^2}{1-(1-\delta)(1+\beta)} + \frac{6(1-\delta)(2-\delta)(1+\frac{1}{\beta})}{(1-(1-\delta)(1+\beta))}\sum_{t=0}^{T-1}(\frac{L^2}{N}\sum_{i=1}^N\mathbb{E}\|x_t^i-\Bar{x}_t\|^2+\mathbb{E}\|\nabla f(\Bar{x}_t)\|^2) \nonumber \\
    &\stackrel{(f)}{\leq}& \frac{8(1-\delta)(2-\delta)T\sigma^2}{\delta^2} + \frac{24(1-\delta)(2-\delta)L^2}{\delta^2 N}\sum_{t=0}^{T-1}\sum_{i=1}^N\mathbb{E}\|x_t^i-\Bar{x}_t\|^2 \nonumber \\
    &&+ \frac{24(1-\delta)(2-\delta)L^2}{\delta^2}\sum_{t=0}^{T-1} \mathbb{E}\|\nabla f(\Bar{x}_t)\|^2 
\end{eqnarray}
\hrulefill
\vspace*{4pt}
\end{figure*}

\begin{figure*}[h]
\normalsize
\setcounter{MYtempeqncnt}{\value{equation}}
\setcounter{equation}{6}
\vspace{-0.2cm}
\begin{eqnarray}\label{theo1_2}
\mathbb{E} [f(\Hat{x}_{t+1})] &\leq& \mathbb{E} [f(\Hat{x}_t)]-\eta \mathbb{E} \langle \nabla f(\Hat{x}_t), \nabla f(\Bar{x}_t) \rangle + \eta \mathbb{E} \langle \nabla f(\Hat{x}_t),\frac{1}{N} \sum_{i=1}^{N} \nabla f_i(\Bar{x}_t)- \nabla f_i(x_t^i) \rangle + \frac{\eta^2 L}{2}\mathbb{E} \|\frac{1}{N} \sum_{i=1}^{N} \nabla f_i(x_t^i,\xi_t^i)\|^2 \nonumber \\
&\leq& \mathbb{E} [f(\Hat{x}_t)]-\frac{\eta}{2} \mathbb{E} (\|\nabla f(\Hat{x}_t)\|^2+\|\nabla f(\Bar{x}_t)\|^2-\|\nabla f(\Hat{x}_t)-\nabla f(\Bar{x}_t)\|^2)\nonumber \\
&&+ \frac{\eta}{2} \mathbb{E} (\|\nabla f(\Hat{x}_t)\|^2+\|\frac{1}{N} \sum_{i=1}^{N} \nabla f_i(\Bar{x}_t)- \nabla f_i(x_t^i)\|^2) + \frac{\eta^2 L}{2}\mathbb{E} \|\frac{1}{N} \sum_{i=1}^{N} \nabla f_i(x_t^i,\xi_t^i)\|^2. 
\end{eqnarray}
\hrulefill
\vspace*{4pt}
\end{figure*}

\setcounter{MYtempeqncnt}{\value{equation}}
\setcounter{equation}{5}
\begin{IEEEproof}[\textbf{Proof of Theorem 1}]
According to the $L$-smoothness of $f(x)$, we have that
\begin{eqnarray}\label{theo1_1}
&&\mathbb{E} [f(\Hat{x}_{t+1})] \nonumber \\
&\leq& \mathbb{E} [f(\Hat{x}_t)]+\mathbb{E} \langle \nabla f(\Hat{x}_t), \Hat{x}_{t+1}-\Hat{x}_t \rangle + \frac{L}{2}\mathbb{E} \|\Hat{x}_{t+1}-\Hat{x}_t\|^2 \nonumber \\
&=& \mathbb{E} [f(\Hat{x}_t)]-\eta \mathbb{E} \langle \nabla f(\Hat{x}_t), \frac{1}{N} \sum_{i=1}^{N} \nabla f_i(x_t^i,\xi_t^i) \rangle  \nonumber \\
&&+ \frac{\eta^2 L}{2}\mathbb{E} \|\frac{1}{N} \sum_{i=1}^{N} \nabla f_i(x_t^i,\xi_t^i)\|^2 \nonumber \\
&\stackrel{(a)}{=}& \mathbb{E} [f(\Hat{x}_t)]-\eta \mathbb{E} \langle \nabla f(\Hat{x}_t), \frac{1}{N} \sum_{i=1}^{N} \nabla f_i(x_t^i) \rangle  \nonumber \\
&&+ \frac{\eta^2 L}{2}\mathbb{E} \|\frac{1}{N} \sum_{i=1}^{N} \nabla f_i(x_t^i,\xi_t^i)\|^2.
\end{eqnarray}
where (a) holds due to $\nabla f_i(x_t^i,\xi_t^i)$ is an unbiased stochastic gradient in expectation. Note that
\begin{eqnarray}
    &&\langle \nabla f(\Hat{x}_t), \frac{1}{N} \sum_{i=1}^{N} \nabla f_i(x_t^i) \rangle \nonumber \\
    &=& \langle \nabla f(\Hat{x}_t), \frac{1}{N} \sum_{i=1}^{N} \nabla f_i(\Bar{x}_t) \rangle \nonumber \\
    &&- \langle \nabla f(\Hat{x}_t),\frac{1}{N} \sum_{i=1}^{N} \nabla f_i(\Bar{x}_t)- \nabla f_i(x_t^i) \rangle \nonumber \\
    &=& \langle \nabla f(\Hat{x}_t), \nabla f(\Bar{x}_t) \rangle \nonumber \\
    &&- \langle \nabla f(\Hat{x}_t),\frac{1}{N} \sum_{i=1}^{N} \nabla f_i(\Bar{x}_t)- \nabla f_i(x_t^i) \rangle. \nonumber
\end{eqnarray}
So we have (\ref{theo1_2}) on the top of next page, where the last inequality comes from the fact that $2\langle a, b \rangle \leq \|a\|^2 + \|b\|^2$. Further, we have
\begin{eqnarray}
&&\mathbb{E} [f(\Hat{x}_{t+1})] \nonumber \\
&\stackrel{(b)}{\leq}&  \mathbb{E} [f(\Hat{x}_t)] - \frac{\eta}{2}\mathbb{E} \|\nabla f(\Bar{x}_t)\|^2 + \frac{\eta L^2}{2}\mathbb{E} \|\Hat{x}_t-\Bar{x}_t\|^2  \nonumber \\
&&+ \frac{\eta L^2}{2N}  \sum_{i=1}^N \mathbb{E} \|\Bar{x}_t-x_t^i\|^2 + \frac{\eta^2 L}{2}\mathbb{E} \|\frac{1}{N} \sum_{i=1}^{N} \nabla f_i(x_t^i,\xi_t^i)\|^2 \nonumber \\
&\stackrel{(c)}{=}&  \mathbb{E} [f(\Hat{x}_t)] - \frac{\eta}{2}\mathbb{E} \|\nabla f(\Bar{x}_t)\|^2 + \frac{\eta^3 L^2}{2}\mathbb{E} \|e_t\|^2 \nonumber \\
&&+ \frac{\eta L^2}{2N}  \sum_{i=1}^N \mathbb{E} \|\Bar{x}_t-x_t^i\|^2 + \frac{\eta^2 L}{2}\mathbb{E} \|\frac{1}{N} \sum_{i=1}^{N} \nabla f_i(x_t^i,\xi_t^i)\|^2. \nonumber
\end{eqnarray}
where (b) follows from Assumption 1 and (c) comes from Lemma~\ref{lemma3}. Summing up the above inequality over $\{0, 1, ..., T-1\}$ yields that
\begin{eqnarray}\label{theo1_4}
    &&\frac{\eta}{2}\sum_{t=0}^{T-1}\mathbb{E} \|\nabla f(\Bar{x}_t)\|^2 \nonumber \\
    &\leq& f(x_0)-f(x^*) +  \frac{\eta L^2}{2N}  \sum_{t=0}^{T-1}  \sum_{i=1}^N \mathbb{E}\|\Bar{x}_t-x_t^i\|^2 \nonumber \\
    &&+ \frac{\eta^3 L^2}{2} \sum_{t=0}^{T-1} \mathbb{E} \|e_t\|^2 + \frac{\eta^2 L}{2}\sum_{t=0}^{T-1} \mathbb{E} \|\frac{1}{N} \sum_{i=1}^{N} \nabla f_i(x_t^i,\xi_t^i)\|^2. \nonumber
\end{eqnarray}
Based on previous lemmas, we further obtain (\ref{theo1_3}) on the next page, where (d) comes from Lemmas \ref{lemma5} and \ref{lemma7}. (e) comes from Lemma \ref{lemma6} and $\eta \leq \frac{\delta}{10L} \leq \frac{\delta}{\sqrt{6}L}$. Consider our assumption $\eta \leq \frac{\delta}{10L} \leq \frac{1}{8L}$, we have that
\begin{eqnarray}
    &&\frac{\eta L^2}{2} + \frac{12(1-\delta)(2-\delta)\eta^3 L^4}{\delta^2} + \eta^2 L^3 \nonumber \\
    &\leq& \frac{\eta L^2}{2} + \frac{\eta L^2}{4} + \frac{\eta L^2}{8} = \frac{7\eta L^2}{8}, \nonumber
\end{eqnarray}
\begin{eqnarray}
    &&\frac{12(1-\delta)(2-\delta)\eta^3 L^2}{\delta^2} + \eta^2 L + \frac{21(1-\delta)\eta^3 L^2}{4\delta^2} \nonumber \\
    &\leq& \frac{\eta}{4} + \eta^2 L + \frac{\eta}{8} = \frac{3\eta}{8} + \eta^2 L, \nonumber
\end{eqnarray}
\begin{equation}
    \frac{4(1-\delta)(2-\delta)\eta^3 L^2}{\delta^2} + \frac{7\eta L^2}{8}\frac{2(1-\delta)\eta^2}{\delta^2} \leq \frac{10 \eta^3 L^2}{\delta^2}. \nonumber
\end{equation}
Combining these results with (\ref{theo1_3}), we further have
\begin{eqnarray}\label{theo3_2}
    &&(\frac{\eta}{8}-\eta^2 L) \sum_{t=0}^{T-1}\mathbb{E}\|\nabla f(\Bar{x}_t)\|^2 \nonumber \\
    &\leq&  f(x_0) - f^* + (\frac{10\eta^3 L^2}{\delta^2}+\frac{\eta^2 L}{2N})T \sigma^2 \nonumber
\end{eqnarray}
Dividing by $\frac{\eta T}{8}$ on both sides of the above inequality yields 
\begin{eqnarray}
    &&\frac{1-8\eta L}{T} \sum_{t=0}^{T-1}\mathbb{E}\|\nabla f(\Bar{x}_t)\|^2 \nonumber \\
    &\leq&  \frac{8(f(x_0) - f^*)}{\eta T} + (\frac{80\eta^2 L^2}{\delta^2}+\frac{4\eta L}{N}) \sigma^2. \nonumber
\end{eqnarray}
\end{IEEEproof}

\begin{figure*}[h]
\normalsize
\setcounter{MYtempeqncnt}{\value{equation}}
\setcounter{equation}{7}
\vspace{-0.2cm}
\begin{eqnarray}\label{theo1_3}
    \frac{\eta}{2} \sum_{t=0}^{T-1}\mathbb{E}\|\nabla f(\Bar{x}_t)\|^2 &\stackrel{(d)}{\leq}& f(x_0) - f^* + \frac{\eta L^2}{2N}\sum_{t=0}^{T-1} \sum_{i=1}^N \mathbb{E}\|\Bar{x}_t-x_t^i\|^2 + \frac{4(1-\delta)(2-\delta)\eta^3 L^2 T \sigma^2}{\delta^2}\nonumber \\
    &&+ \frac{12(1-\delta)(2-\delta)\eta^3 L^4}{\delta^2 N}\sum_{t=0}^{T-1}\sum_{i=1}^N \mathbb{E}\|\Bar{x}_t-x_t^i\|^2 + \frac{12(1-\delta)(2-\delta)\eta^3 L^2}{\delta^2}\sum_{t=0}^{T-1} \mathbb{E}\|\nabla f(\Bar{x}_t)\|^2  \nonumber \\
    &&+\frac{\eta^2 L}{2}(\frac{\sigma^2 T}{N}+ \frac{2L^2}{N}\sum_{t=0}^{T-1} \sum_{i=1}^N \mathbb{E}\|\Bar{x}_t-x_t^i\|^2 + 2\sum_{t=0}^{T-1}\mathbb{E}\|\nabla f(\Bar{x}_t)\|^2) \nonumber \\
    &\stackrel{(e)}{\leq}& f(x_0) - f^* + (\frac{4(1-\delta)(2-\delta)\eta^3 L^2}{\delta^2}+\frac{\eta^2 L}{2N})T\sigma^2 + (\frac{12(1-\delta)(2-\delta)\eta^3 L^2}{\delta^2}+\eta^2 L)\sum_{t=0}^{T-1}\mathbb{E}\|\nabla f(\Bar{x}_t)\|^2 \nonumber \\
    &&+ (\frac{\eta L^2}{2} + \frac{12(1-\delta)(2-\delta)\eta^3 L^4}{\delta^2} + \eta^2 L^3)(\frac{2(1-\delta)\eta^2}{\delta^2}T \sigma^2 + \frac{6(1-\delta)\eta^2}{\delta^2}\sum_{t=0}^{T-1} \mathbb{E}\|\nabla f(\Bar{x}_t)\|^2). 
\end{eqnarray}
\hrulefill
\vspace*{4pt}
\end{figure*}

\begin{IEEEproof}[\textbf{Proof of Corollary 1}]
Consider that we set
\setcounter{equation}{8}
\begin{equation}
    \eta = \frac{1}{\sqrt{\frac{T}{N}}+L+\frac{T^{1/3}}{\delta^{2/3}}},
\end{equation}
then we have
\begin{equation}
    \frac{80\eta^2 L^2 \sigma^2}{\delta^2} \leq \frac{80 L^2 \sigma^2}{\delta^2}\frac{1}{(\frac{T^{1/3}}{\delta^{2/3}})^2} = O\bigg(\frac{1}{\delta^{2/3} T^{2/3}}\bigg). \nonumber
\end{equation}
and
\begin{equation}
    \frac{4\eta L \sigma^2}{N} \leq \frac{4L \sigma^2}{N\sqrt{\frac{T}{N}}} = O\bigg(\frac{1}{\sqrt{NT}}\bigg).
\end{equation}
Finally, for the first term, we have
\begin{eqnarray}
    &&\frac{8(f(x_0)-f(x^*))}{\eta T} \nonumber \\
    &=& \frac{8(f(x_0)-f(x^*))(\sqrt{\frac{T}{N}}+L+\frac{T^{1/3}}{\delta^{2/3}})}{T} \nonumber \\
    &=& O\bigg(\frac{1}{\sqrt{NT}}+\frac{1}{T}+\frac{1}{\delta^{2/3} T^{2/3}}\bigg). \nonumber
\end{eqnarray}
Thus we complete the proof.
\end{IEEEproof}

\begin{IEEEproof}[\textbf{Proof of Lemma 1}]
Similar to the proof of Lemma \ref{lemma7}, we have that
\begin{eqnarray}
\mathbb{E}\|e_t\|^2 &\leq& (1-\delta)(1+1/\beta) \mathbb{E}\|\frac{1}{N} \sum_{i=1}^N p_{t-1}^i\|^2 \nonumber \\
&&+ (1-\delta)(1+\beta) \mathbb{E}\|e_{t-1}\|^2 \nonumber \\
&=& \frac{(1-\delta)(1+1/\beta)}{N^2}  \mathbb{E}\|\sum_{i=1}^N \mathcal{C}_\delta (\nabla f_i(x_{t-1}^i,\xi_{t-1}^i))\|^2 \nonumber \\
&&+ (1-\delta)(1+\beta) \mathbb{E}\|e_{t-1}\|^2. \nonumber
\end{eqnarray}
Further, we have (\ref{lemma1_1}) on the next page, where (a) comes from Assumption 4. (b) and (c) follow from Assumption 3. (d) holds by setting $\beta = \frac{\delta}{2(1-\delta)}$.
\end{IEEEproof}

\begin{figure*}[h]
\normalsize
\setcounter{MYtempeqncnt}{\value{equation}}
\setcounter{equation}{10}
\vspace{-0.1cm}
\begin{eqnarray}\label{lemma1_1}
\mathbb{E}\|e_t\|^2 &=& (1-\delta)(1+\beta) \mathbb{E}\|e_{t-1}\|^2 + \frac{(1-\delta)(1+1/\beta)}{N^2}\bigg(\sum_{i=1}^N \mathbb{E}\|\mathcal{C}_\delta (\nabla f_i(x_{t-1}^i,\xi_{t-1}^i))\|^2 \nonumber \\
&&+ \sum_{i \neq j} \mathbb{E} \langle \mathcal{C}_\delta (\nabla f_i(x_{t-1}^i,\xi_{t-1}^i)), \mathcal{C}_\delta (\nabla f_j(x_{t-1}^j,\xi_{t-1}^j)) \rangle \bigg)\nonumber \\
&\stackrel{(a)}{=}& (1-\delta)(1+\beta) \mathbb{E}\|e_{t-1}\|^2 + \frac{(1-\delta)(1+1/\beta)}{N^2}\bigg(\sum_{i=1}^N \mathbb{E}\|\mathcal{C}_\delta(\nabla f_i(x_{t-1}^i,\xi_{t-1}^i)) \nonumber \\
&&-\nabla f_i(x_{t-1}^i,\xi_{t-1}^i)+\nabla f_i(x_{t-1}^i,\xi_{t-1}^i)\|^2+ \delta^2\sum_{i \neq j} \mathbb{E} \langle \nabla f_i(x_{t-1}^i,\xi_{t-1}^i), \nabla f_j(x_{t-1}^j,\xi_{t-1}^j) \rangle \bigg)\nonumber \\
&\stackrel{(b)}{\leq}& (1-\delta)(1+\beta) \mathbb{E}\|e_{t-1}\|^2 + \frac{(1-\delta)(1+1/\beta)}{N^2}\bigg(2(2-\delta)N M^2 \nonumber \\
&&+ \delta^2\sum_{i \neq j} \mathbb{E} \|\nabla f_i(x_{t-1}^i,\xi_{t-1}^i)\| \mathbb{E}\|\nabla f_j(x_{t-1}^j,\xi_{t-1}^j)\| \bigg)\nonumber \\
&\stackrel{(c)}{\leq}& (1-\delta)(1+\beta) \mathbb{E}\|e_{t-1}\|^2 + \frac{(1-\delta)(1+1/\beta)}{N^2}\bigg(2(2-\delta)N M^2+ \delta^2 N (N-1) M^2 \bigg)\nonumber \\
&\leq& \frac{(1-\delta)(1+1/\beta)(2(2-\delta)+\delta^2 (N-1))M^2}{(1-(1-\delta)(1+\beta))N} \stackrel{(d)}{\leq} \frac{4(1-\delta)(2(2-\delta)+\delta^2 (N-1))M^2}{\delta^2 N}. 
\end{eqnarray}
\hrulefill
\vspace*{4pt}
\end{figure*}

\newpage

\begin{IEEEproof}[\textbf{Proof of Lemma 2}]
Similar to the proof of Lemma \ref{lemma6}, we could obtain (\ref{lemma2_1}) on the next page, where (a) comes from the Definition 1 and Assumption 4. (b) follows from Assumption 3. 
Thus, we have
\setcounter{equation}{12}
\begin{eqnarray}
&&\frac{1}{N}\sum_{i=1}^N \mathbb{E}\|\Bar{x}_t-x_t^i\|^2\nonumber \\
&\leq& \eta^2 M^2( (1-\delta)(t-t')  + (1-\delta)^2(t-t')(t-t'-1) ) \nonumber \\
&\stackrel{(c)}{\leq}& \frac{(1-\delta)(1-\delta+\delta^2)}{\delta^2}\eta^2 M^2.
\end{eqnarray}
(c) holds because $t-t' \leq \lfloor \frac{1}{\delta} \rfloor \leq \frac{1}{\delta}$.

\end{IEEEproof}

\begin{figure*}[h]
\normalsize
\setcounter{MYtempeqncnt}{\value{equation}}
\setcounter{equation}{11}
\begin{eqnarray}\label{lemma2_1}
\sum_{i=1}^N \mathbb{E}\|\Bar{x}_t-x_t^i\|^2
&\leq& \eta^2 \sum_{i=1}^N \mathbb{E}\|\sum_{\tau=t'}^{t-1}\nabla f_i(x_\tau^i,\xi_\tau^i)-\mathcal{C}_\delta(\nabla f_i(x_\tau^i,\xi_\tau^i))\|^2.\nonumber \nonumber \\
&=& \eta^2 \sum_{i=1}^N \bigg( \sum_{\tau=t'}^{t-1} \mathbb{E}\|\nabla f_i(x_\tau^i,\xi_\tau^i)-\mathcal{C}_\delta(\nabla f_i(x_\tau^i,\xi_\tau^i))\|^2 \nonumber \\
&&+ \sum_{p \neq q} \mathbb{E} \langle \nabla f_i(x_p^i,\xi_p^i)-\mathcal{C}_\delta(\nabla f_i(x_p^i,\xi_p^i)), \nabla f_i(x_q^i,\xi_q^i)-\mathcal{C}_\delta(\nabla f_i(x_q^i,\xi_q^i)) \rangle \bigg) \nonumber \\
&\stackrel{(a)}{\leq}& \eta^2 \sum_{i=1}^N \bigg( (1-\delta)\sum_{\tau=t'}^{t-1} \mathbb{E}\|\nabla f_i(x_\tau^i,\xi_\tau^i)\|^2 + (1-\delta)^2 \sum_{p \neq q} \mathbb{E} \langle \nabla f_i(x_p^i,\xi_p^i), \nabla f_i(x_q^i,\xi_q^i) \rangle \bigg) \nonumber \\ 
&\stackrel{(b)}{\leq}& \eta^2 \sum_{i=1}^N \bigg( (1-\delta)(t-t') M^2 + (1-\delta)^2(t-t')(t-t'-1)M^2 \bigg). 
\end{eqnarray}

\hrulefill
\vspace*{4pt}
\end{figure*}


\begin{IEEEproof}[\textbf{Proof of Theorem 2}]
Similar to the proof of Theorem 1, according to the $L$-smoothness of $f(x)$, we have that
\setcounter{equation}{13}
\begin{eqnarray}\label{theo2_1}
    &&\frac{\eta}{2}\sum_{t=0}^{T-1}\mathbb{E} \|\nabla f(\Bar{x}_t)\|^2 \nonumber \\ 
    &\leq& f(x_0)-f(x^*) + \frac{\eta^2 L}{2}\sum_{t=0}^{T-1} \mathbb{E} \|\frac{1}{N} \sum_{i=1}^{N} \nabla f_i(x_t^i,\xi_t^i)\|^2 \nonumber \\
    &&+ \frac{\eta^3 L^2}{2} \sum_{t=0}^{T-1} \mathbb{E} \|e_t\|^2 + \frac{\eta L^2}{2N}  \sum_{t=0}^{T-1}  \sum_{i=1}^N \mathbb{E}\|\Bar{x}_t-x_t^i\|^2. 
\end{eqnarray}
Substituting Lemmas 1, 2 and 5 into (\ref{theo2_1}) yields that

\begin{eqnarray}
&&\frac{\eta}{2}\sum_{t=0}^{T-1} \mathbb{E}\|\nabla f(\Bar{x}_t)\|^2 \nonumber \\
&\leq& f(x_0)-f(x^*) \nonumber \\
&&+\frac{2(1-\delta)(2(2-\delta)+\delta^2 (N-1))\eta^3 L^2 M^2 T}{\delta^2 N} \nonumber \\
&&+ (\frac{\eta L^2}{2}+\eta^2 L^3)\frac{(1-\delta)(1-\delta+\delta^2)\eta^2 M^2 T}{\delta^2} \nonumber \\
&&+ \frac{\eta^2 L T \sigma^2}{2N} + \eta^2 L \sum_{t=0}^{T-1} \|\nabla f(\Bar{x}_t)\|^2. \nonumber
\end{eqnarray}
Dividing by $\frac{\eta T}{2}$ on both sides of the above inequality and consider that $\eta L < \frac{1}{2}$, we obtain the final result.
\end{IEEEproof}

\begin{IEEEproof}[\textbf{Proof of Theorem 3}]
We follow the proof of theorem 1. Based on Assumption 5, we modify Lemma \ref{lemma4} as
\begin{eqnarray}
    &&\mathbb{E}\|\nabla f_i(x_t^i, \xi_t^i)\|^2 \nonumber \\
    &\leq& \sigma^2 + \mathbb{E}\|\nabla f_i(x_t^i) - \nabla f_i(\Bar{x}_t) + \nabla f_i(\Bar{x}_t)\|^2 \nonumber \\
    &\leq& \sigma^2 + 2\mathbb{E}\|\nabla f_i(x_t^i) - \nabla f_i(\Bar{x}_t)\|^2 + 2\mathbb{E}\|\nabla f_i(\Bar{x}_t)\|^2 \nonumber \\
    &\leq& \sigma^2 + 2L^2 \mathbb{E}\|\Bar{x}_t-x_t^i\|^2 + 2\mathbb{E}\|\nabla f_i(\Bar{x}_t)\|^2 \nonumber \\
    &\leq& \sigma^2 + 2L^2 \mathbb{E}\|\Bar{x}_t-x_t^i\|^2 + 2\rho\mathbb{E}\|\nabla f(\Bar{x}_t)\|^2,
\end{eqnarray}
where the last inequality comes from Assumption 5. This result is different from Lemma 4 only in the constant coefficient. The remaining proof details could follow the proof of Theorem 1. Thus, the final convergence rate is in the same order as Theorem 1 and Corollary 1. We ignore the detailed formula here.
\end{IEEEproof}

\begin{figure*}[!t]
\centering
\includegraphics[scale=0.29]{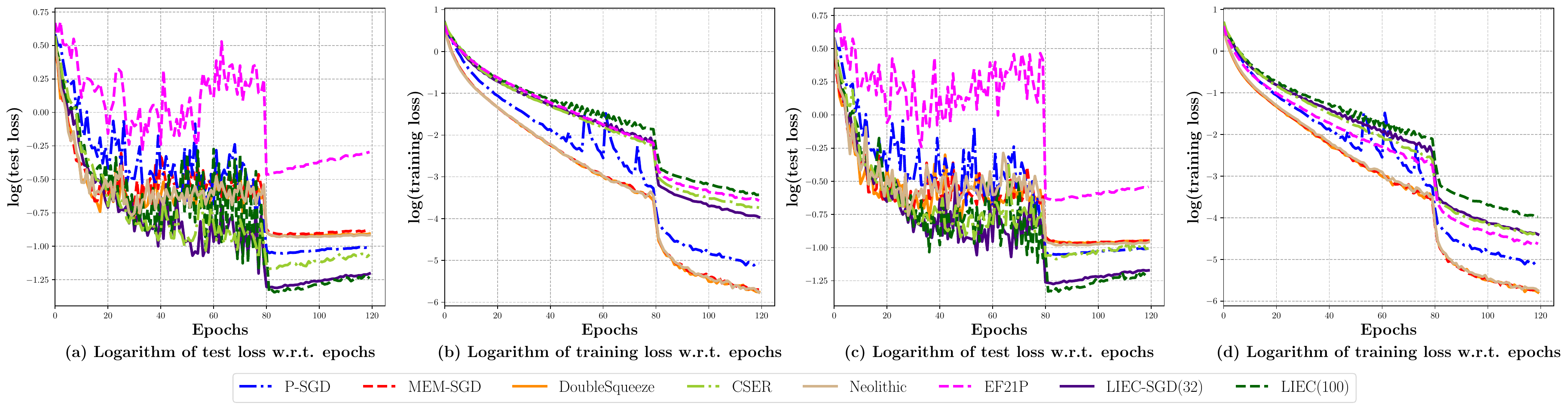}
\vspace{-0.2cm}
\caption{Logarithm of training or test loss of training ResNet18 on CIFAR-10 w.r.t. epochs. (a): Test loss when adopting SignSGD. (b): Training loss when adopting SignSGD. (c): Test loss when adopting Blockwise-SignSGD. (d): Training loss when adopting Blockwise-SignSGD.}
\label{cifar10_loss}
\end{figure*}

\begin{figure*}[!t]
\centering
\includegraphics[scale=0.29]{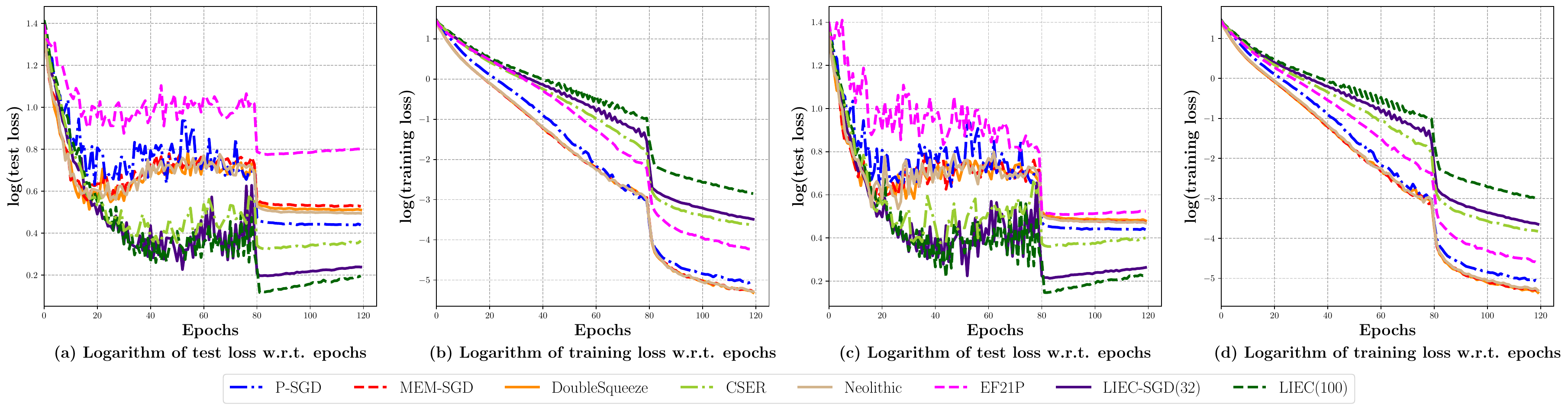}
\vspace{-0.2cm}
\caption{Logarithm of training or test loss of training ResNet34 on CIFAR-100 w.r.t. epochs. (a): Test loss when adopting SignSGD. (b): Training loss when adopting SignSGD. (c): Test loss when adopting Blockwise-SignSGD. (d): Training loss when adopting Blockwise-SignSGD.}
\label{cifar100_loss}
\end{figure*}

\newpage

\section{More Experimental Results}

In this section, we provide more experimental results. We show the loss value w.r.t. epochs of training ResNet18 on CIFAR-10 and training ResNet34 on CIFAR-100 in Fig. \ref{cifar10_loss} and \ref{cifar100_loss} respectively. 

In the subfigures (a) and (c), we plot the logarithm of test loss with different compression operators. We can find that LIEC-SGD converges to the lowest value and is well ahead of other baselines. The curves of logarithm of training loss are shown in subfigures (b) and (d). MEM-SGD and DoubleSqueeze show advantages in this indicator. Here, we point out that the test loss is evaluated on the test data under the same distribution as training data, which corresponds to optimizing a sum of stochastic functions: $f(x) = \frac{1}{N} \sum_{i=1}^N \mathbb{E}_{\xi_i \sim \mathcal{D}_i} f_i(x,\xi_i)$. On the other hand, the training loss is evaluated on the finite training set, which corresponds to optimizing a finite-sum function: $f(x) = \frac{1}{N} \sum_{i=1}^N \sum_{\xi_j \in \mathcal{S}_i} f_i(x,\xi_j)$, where $\mathcal{S}_i$ represents the training set on the $i$-th worker. From the optimization perspective, since our goal is to optimize the former function, subfigures (a) and (c) which plot the test loss represents the metric we mainly target on.

%

%




\end{document}